\documentclass[10pt,journal,compsoc]{IEEEtran}
\usepackage{times}
\usepackage{textcomp}
\usepackage{stfloats}
\usepackage{ragged2e}
\usepackage{url}
\usepackage{verbatim}
\usepackage{graphicx}
\usepackage[marginal]{footmisc}
\usepackage{epsfig}
\usepackage{amssymb}
\usepackage{multirow}
\usepackage{multicol}
\usepackage{booktabs}
\graphicspath{{figures/}}
\usepackage{ragged2e}
\usepackage{amsmath}
\usepackage[pagebackref=true,breaklinks,colorlinks,citecolor=blue,linkcolor=blue,urlcolor=blue,hypertexnames=false]{hyperref}
\usepackage{cleveref}
\usepackage{marvosym}
\usepackage{subcaption}
\usepackage{caption}
\usepackage{longtable}
\usepackage{sidecap}

\usepackage{color}
\usepackage{colortbl}%
\newcommand{\myrowcolour}{\rowcolor[gray]{0.925}}

\usepackage{amsmath,amsfonts}
\usepackage{array}
\def\BibTeX{{\rm B\kern-.05em{\sc i\kern-.025em b}\kern-.08em
    T\kern-.1667em\lower.7ex\hbox{E}\kern-.125emX}}
\usepackage{balance}

\usepackage{algorithm,algpseudocode}
\usepackage{xcolor}
\usepackage{tikz}
\usetikzlibrary{fit,calc}

\usepackage{xr}
\makeatletter

\newcounter{algsubstate}

\newcommand*{\tikzmk}[1]{\tikz[remember picture,overlay,] \node[inner sep=0mm,outer sep=0mm] (#1) {};\ignorespaces}
\newcommand{\boxit}[1]{\tikz[remember picture,overlay]{\node[inner sep=0mm,outer sep=0mm,xshift=-36pt,fill=#1,opacity=.25,fit={(A)($(B)+(1.10\linewidth,-0.1\baselineskip)$)}] {};}\ignorespaces}
\colorlet{pink}{red!40}
\colorlet{blue2}{cyan!60}
\colorlet{green}{green!40}

\begin{document}
\title{Re-boosting Self-Collaboration Parallel Prompt GAN for Unsupervised Image Restoration}
\author{Xin Lin\textsuperscript{$\ast$},
Yuyan Zhou\textsuperscript{$\ast$},
Jingtong Yue,
Chao Ren\textsuperscript{\Letter}, 
Kelvin C.K. Chan,
Lu Qi,
Ming-Hsuan Yang
\IEEEcompsocitemizethanks{
\IEEEcompsocthanksitem This work was supported in part by the National Natural Science Foundation of China under Grant 62171304 and partly by the Natural Science Foundation of Sichuan Province under Grant 2024NSFSC1423 and the TCL Science and Technology Innovation Fund under Grant 25JZH008, and the Young Faculty Technology Innovation Capacity Enhancement Program of Sichuan University under Grant 2024SCUQJTX025.
\IEEEcompsocthanksitem $*$ denotes equal contributions, \Letter\, denotes corresponding author.
\IEEEcompsocthanksitem Xin Lin, Jingtong Yue, and Chao Ren are with the College of Electronics and Information Engineering, Sichuan University, Chengdu 610065, China. E-mails: 
linxin020826@gmail.com; yuejingtong@stu.scu.edu.cn; chaoren@scu.edu.cn.
\IEEEcompsocthanksitem Lu Qi and Ming-Hsuan Yang are with the University of California at Merced, Merced, CA 95343 USA. Emails: \{lqi5, mhyang\}@ucmerced.edu.
\IEEEcompsocthanksitem Yuyan Zhou is with with the Department of Computer Science
and Engineering, Hong Kong University of Science and Technology, Hong
Kong. Email: 0903yyz@gmail.com.
\IEEEcompsocthanksitem Kelvin C.K. Chan is with Google DeepMind, USA. Email: kelvinckchan@google.com.}
}

\IEEEtitleabstractindextext{
\begin{abstract}
\justifying
Deep learning methods have demonstrated state-of-the-art performance in image restoration, especially when trained on large-scale paired datasets. 
However, acquiring paired data in real-world scenarios poses a significant challenge.
Unsupervised restoration approaches based on generative adversarial networks (GANs) offer a promising solution without requiring paired datasets. 
Yet, these GAN-based approaches struggle to surpass the performance of conventional unsupervised GAN-based frameworks without significantly modifying model structures or increasing the computational complexity.
To address these issues, we propose a self-collaboration (SC) strategy for existing restoration models. 
This strategy utilizes information from the previous stage as feedback to guide subsequent stages, achieving significant performance improvement without increasing the framework's inference complexity. 
The SC strategy comprises a prompt learning (PL) module and a restorer ($Res$).
It iteratively replaces the previous less powerful fixed restorer $\overline{Res}$ in the PL module with a more powerful $Res$.
The enhanced PL module generates better pseudo-degraded/clean image pairs, leading to a more powerful $Res$ for the next iteration. Our SC can significantly improve the $Res$'s performance by over 1.5 dB without adding extra parameters or computational complexity during inference.
Meanwhile, existing self-ensemble (SE) and our SC strategies enhance the performance of pre-trained restorers from different perspectives. 
As SE increases computational complexity during inference, we propose a re-boosting module to the SC (Reb-SC) to improve the SC strategy further by incorporating SE into SC without increasing inference time. 
This approach further enhances the restorer's performance by approximately 0.3 dB.
Additionally, we present a baseline framework that includes parallel generative adversarial branches with complementary ``self-synthesis'' and ``unpaired-synthesis'' constraints, ensuring the effectiveness of the training framework. 
Extensive experimental results on restoration tasks demonstrate that the proposed model performs favorably against existing state-of-the-art unsupervised restoration methods.
Source code and trained models are publicly available at: \url{https://github.com/linxin0/RSCP2GAN}.
\end{abstract}

}

\maketitle
\IEEEdisplaynontitleabstractindextext

\IEEEpeerreviewmaketitle
\IEEEraisesectionheading{\section{Introduction}\label{intro}}

\IEEEPARstart{I}{mage} restoration aims to recover high-quality, visually pleasing images from degraded observations, which is a classical problem in computer vision. 
Early methods leverage physical priors to constrain the solution space and recover latent clean images \cite{chuantong1, chuantong2, chuantong3, chuantong4}. 
However, these methods, constrained by empirical statistical priors, often struggle with the complexity and variability of real-world degraded images, leading to unreliable results. 
With the advances of deep learning, recent learning-based methods achieve state-of-the-art results by training deep neural networks on paired degraded/clean datasets using supervised learning \cite{naf, restormer, aind1, aind2, dehazing1, dehazing5, fxy1, rcd, pan_1, pan_2, pan_3}.
However, the lack of paired training data is one of the biggest obstacles in these tasks. 
Creating a large amount of paired training data is time-consuming and labor-intensive.

To address this issue, unsupervised image restoration methods \cite{chen, hong, dbsnl, learn, deraincyclegan, pjs} have emerged, leveraging the generative adversarial network (GAN) framework. 
These methods aim to generate high-quality pseudo-degraded images to train effective restoration models (restorers). 
However, the performance of restorers trained with current unsupervised frameworks is limited. 
As noted in GAN2GAN \cite{cha}, a primary limitation is the gap between real and pseudo degraded images, and thus a model using multiple generators and discriminators is proposed to generate images to better match the real noise distribution. 
Furthermore, existing frameworks cannot improve the restoration potential without significantly changing their structure or increasing the inference complexity (e.g., using a certain self-ensemble strategy).

\begin{figure*}[t]
\includegraphics[width=1\linewidth]{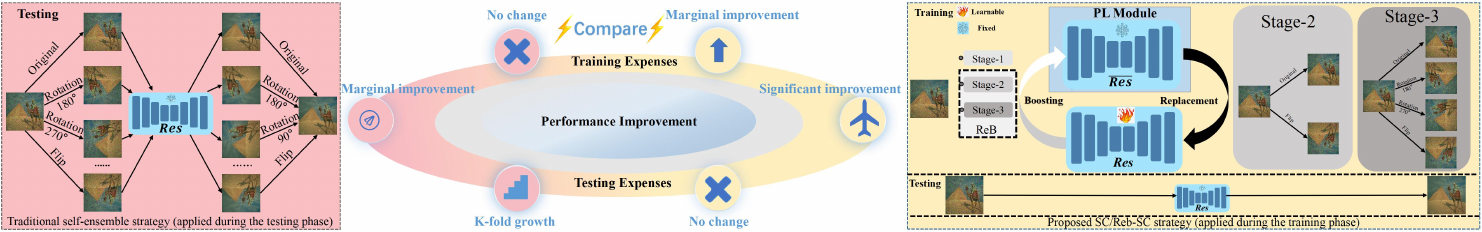}
\caption{Comparison of the proposed self-collaboration (SC) and re-boosting SC (Reb-SC) strategies with the conventional self-ensemble (SE) strategy in training expenses, testing expenses, and performance improvement, respectively.}
\label{diyige}
\end{figure*}

To address the above-mentioned issues, we introduce an innovative unsupervised restoration framework called Re-boosting Self Collaboration Parallel Prompt GAN (RSCP$^{2}$GAN). 
The core self-collaboration (SC) strategy provides the framework with an effective self-boosting capability, enabling the restorer obtained from the conventional GAN framework to evolve continuously and significantly.
Specifically, it consists of a prompt learning (PL) module and a restorer ($Res$). 
The SC strategy operates iteratively by replacing the previous, less powerful fixed restorer $\overline{Res}$ in the PL module with the current, more capable $Res$. 
The updated PL module then generates higher-quality pseudo-degraded images, further enhancing the $Res$ in subsequent iterations. 
The comparative analysis with the conventional self-ensemble (SE) strategy is shown in Fig. \ref{diyige}. Traditional SE is applied during the testing phase of a trained model, where the input is augmented, followed by multi-branch restoration and averaging. This significantly increases both the testing time and memory usage, which is disadvantageous when applying restoration networks in the real world. Meanwhile, it obtains only a marginal improvement. In contrast, our SC can significantly improve the restorer's ($Res$) performance by over 1.5 dB without incurring extra testing expenses.

Both SE and SC strategies aim to improve $Res$'s performance but from different perspectives: SE focuses on data augmentation during inference, while SC enhances cooperation between the restorer and the generator during training. 
Building on this, we propose a re-boosting module for SC, termed the Reb-SC strategy. The re-boosting applies data augmentation during the training phase in our SC strategy, further enhancing the proposed SC. The Reb-SC, like the original SC, is applied during the training phase, improving the performance of the restorer significantly and making only minor changes to the framework, without affecting testing time or memory usage, and having no impact on the application of restoration networks. By this way, our SC can further improve the performance of the restorer. Please see Sec. \ref{analysis_rebsc} for more detailed analysis.
Extensive experimental results on denoising, deraining, and desnowing tasks confirm the superiority of our method.

The main contributions of this work are:
\begin{itemize}
\item[$\bullet$] We propose a Self-Collaboration (SC) strategy that significantly enhances the performance of GAN-based restoration frameworks without increasing inference computational complexity. 
This strategy relies on two key components: the prompt learning (PL) module and the restorer ($Res$). 
The PL module is the core part of the prompt-guided degraded image generator, capable of synthesizing 
high-quality degraded images.
\item[$\bullet$] We introduce a parallel prompt GAN framework that incorporates complementary ``self-synthesis'' and ``unpaired-synthesis'' constraints, serving as a robust baseline for image restoration.
\item[$\bullet$] We present a re-boosting module to increase further the effectiveness of the proposed SC strategy, i.e., Reb-SC. Similar to the SC strategy, the inference time also does not increase, and the performance of the restorer ($Res$) can be enhanced further.
\item[$\bullet$] We conduct comprehensive experiments on restoration tasks, demonstrating that our RSCP$^{2}$GAN achieves strong performance across various datasets.
\end{itemize}

This work significantly extends our prior work \cite{scpgabnet}.
The key advancements in this paper include:
1) Extension to General Restoration Framework: Our early work focuses only on image denoising tasks. 
This paper extends the previous denoising model and noise extract (NE) module to the restoration framework and degradation prompt learning (PL) module, resulting in the generalized Self-Collaboration Parallel Prompt GAN (SCP$^{2}$GAN) framework.
2) Introduction of Reb-SC Strategy: We introduce the Reb-SC strategy to further improve the performance of the PL module without additional parameters and further improve the performance of the restorer. 
The framework in this work is RSCP$^{2}$GAN.
3) Extended Experiments and Analysis: We perform extensive experiments on multiple datasets, including deraining (Rain100L \cite{rain100}, Rain12 \cite{rain12}, RealRainL \cite{realrain}), desnowing (CSD \cite{csd} and Snow100K \cite{snow100k}) and denoising (SIDD \cite{sidd}, DND \cite{dnd}, PolyU \cite{polyu}), and our method performs well than state-of-the-art techniques, demonstrating RSCP$^{2}$GAN's superior performance.
4) Analysis of Training/Testing Expenses and Strategy Effectiveness: We add more analyses about training expenses, testing expenses, and performance improvement on the proposed SC\&Reb-SC strategies and traditional self-ensemble augmentation to clarify our contribution.

\section{Related Work}

\subsection{Supervised Image Restoration}
In recent years, supervised data-driven CNN models have been shown to outperform conventional image restoration methods in various tasks such as image denoising \cite{17, 54, ridnet, aind1, aind2}, image deraining \cite{sgi, lrlrnet, rcd, drsformer, rlp}, image dehazing \cite{dehazing1, dehazing2, dehazing3, dehazing4, dehazing5}, and image declaring \cite{zhou2023improving, wu2021train, dai2022flare7k}.
These approaches typically involve designing effective restorers trained using pairs of clean and degraded image datasets captured from real scenes.

{\flushleft \textbf{Image denoising.}} 
The RIDNet method \cite{38} combines synthetic and real images during training to enhance the model's generality for denoising. 
On the other hand, Cheng et al. \cite{99} generate a set of image basis vectors from the noisy input images and reconstruct them from the subspace formed by these basis vectors to obtain image-denoising results.
Numerous approaches simultaneously address Gaussian and real-world noise \cite{aind1, aind2}.
NAFNet \cite{naf} incorporates a series of straightforward but highly effective enhancements, refining the network and fully realizing its performance potential.
Recently, a transformer-based framework \cite{restormer} has been developed, leveraging the advantages of the self-attention strategy while reducing computational complexity.
{\flushleft \textbf{Image deraining and desnowing.}} 
As common forms of weather-induced degradation, rain and snow have received considerable attention in image restoration. To tackle this challenge, numerous studies have introduced learning-based approaches that demonstrate strong performance in restoring images degraded by rain or snow. For image deraining, Li et al. \cite{realrain} use real-world rainy video clips to establish a high-quality dataset named RealRainL, consisting of 1,120 high-resolution paired clean and rainy images with low- and high-density rain streaks.
On the other hand, SGINet \cite{sgi} uses high-level semantic information to improve rain removal and
RadNet \cite{78} simultaneously removes rain streaks and raindrops.
Recently, MIRNet \cite{mir} presents an effective feature extraction module to facilitate image restoration and enhancement, and 
RCDNet \cite{rcd} proposes an unfolding technique, employing multi-stage training with M-net and B-net to achieve better deraining results. 
Most recently, NAFNet \cite{naf} shows a simple yet effective and efficient approach that achieves state-of-the-art performance for image restoration with using nonlinear activation functions. 
Restormer \cite{restormer} utilizes the attention mechanism for deraining and leverages the transformer framework for deraining and optimizing their benefits while reducing computational complexity. 
A multi-expert-based DRSFormer \cite{drsformer} provides more accurate detail and texture recovery. 
For image desnowing, Zhang et al. \cite{snow_1} present a densely connected multi-scale architecture that leverages semantic segmentation and depth priors via a self-attention mechanism to enhance image restoration performance. Chen et al. \cite{csd} propose a single-image desnowing framework that integrates a hierarchical dual-tree complex wavelet transform for better representation of snow particles of various shapes and scales. They further introduced a novel, contrast channel loss to exploit the intensity differences between snowy and clean regions, significantly improving snow removal performance. 
Quan et al. \cite{quan2023image} propose InvDSNet, a dual-path invertible neural network for single image desnowing, which enables effective snowflake removal while preserving image details through progressive feature disentanglement and invertible reconstruction.
Meanwhile, PromptIR \cite{potlapalli2023promptir} presents a prompt-learning-based framework for all-in-one blind image restoration, aiming to handle various degradation types, including deraining, within a unified model. The method introduces a lightweight Prompt Block module that generates input-conditioned prompts to encode degradation-specific cues, which are then injected into multiple stages of the transformer decoder to guide the restoration process.
After that, LMQFormer \cite{lmqformer}, a lightweight yet effective snow removal network, is proposed, which leverages a Laplace-guided VQVAE to generate a coarse snow mask and a Mask Query Transformer to refine the restoration. By introducing duplicated mask query attention, the model focuses computation on snow regions, achieving state-of-the-art performance with minimal parameters and fast inference speed. 
To better model the spatially-varying and multi-scale nature of rain streaks, Chen et al. \cite{nerd} develop a bidirectional multi-scale Transformer architecture augmented with implicit neural representations (INRs). Unlike previous single-scale Transformer-based methods that process rain effects at a fixed resolution, their model leverages both coarse-to-fine and fine-to-coarse feature propagation to exploit cross-scale dependencies.
More recently, PEUNet \cite{peunet} incorporates physical priors—such as atmospheric light and snow shape—to refine the desnowing process iteratively. By formulating snow removal as an optimization problem under a MAP framework and introducing a novel snow shape prior as surrogate supervision, their method enhances both interpretability and performance, achieving state-of-the-art results across multiple benchmarks.

However, the number of real-world degraded images is limited, and creating large amounts of paired training data is time-consuming and labor-intensive. 

\subsection{Unsupervised Image Restoration}
In unsupervised settings, due to the lack of real paired datasets, synthetic images are commonly used to approximate real training data. Thus, it is crucial to address the domain gap between the synthetic and real image domains. Fu et al. \cite{re_1} introduce a CNN-based reconstruction method leveraging both external priors and image-specific internal learning, demonstrating strong performance and generalization on real-world coded hyperspectral data. Da et al. \cite{re_2} propose PromptGAT, a novel sim-to-real transfer framework that leverages prompt-based learning with large language models to enhance the adaptability of reinforcement learning policies to the real-world traffic signal control task. Meanwhile, numerous unsupervised restoration methods have been developed using a large amount of synthetic data from generative adversarial networks to train the model. In the following sections, we review how prior works tackle the domain gap and enhance the performance of restoration networks under unsupervised settings in image denoising, deraining, and desnowing tasks.

{\flushleft \textbf{Image denoising.}} GCBD \cite{chen} uses a generator capable of producing pseudo-noisy images to train a denoiser, and CycleGAN \cite{patch} is introduced for further improvement. 
Among these methods, GAN2GAN \cite{cha} uses a multi-generator / discriminator architecture to enhance the extraction of noisy information and generate synthetic images that closely match the real noise distribution. 
On the other hand, Hong et al. \cite{hong} introduce UIDNet, which utilizes a sharpening processing mechanism to achieve noise separation and improve the training of unpaired denoising models. 
In SCPGabNet \cite{scpgabnet}, a self-collaboration strategy that iteratively enhances the performance of the denoising network has been proposed, leading to significant improvements over conventional GAN frameworks. 
Additionally, various methods have emerged for training models exclusively with noisy images, called self-supervised denoising. CVF-SID \cite{cvf} integrates cyclic adversarial learning with the self-supervised residual framework.
Recently, a self-supervised framework named AP-BSN \cite{apbsn} has been shown to effectively manage real-world signal-dependent noise and adapt well to realistic noise conditions. 
Most recently, LG-BPN \cite{lgbpn} shows masking the central region of a large convolution kernel to reduce the spatial correlation of noise and introduces a dilated Transformer block to capture global information, while others are developed for APBSN using random sampling for augmentation \cite{pyz}.

{\flushleft \textbf{Image deraining and desnowing.}} Numerous unsupervised image restoration methods have been developed \cite{unrain1, unrain2, unrain3, deraincyclegan} based on the CycleGAN model. 
The DerainCycleGAN \cite{deraincyclegan} extracts the rain streak masks using two constrained cycle-consistency branches by paying attention to both the rainy and rain-free image domains for restoration. 
Yu et al. \cite{96} consider the prior knowledge of the rain streak and connect the model-driven and data-driven methods via an unsupervised learning framework.
Yasarla et al. \cite{yasarla2022unsupervised} introduce a weather-agnostic unsupervised restoration approach by augmenting CycleGAN with Deep Gaussian Process-based latent supervision, enabling effective training on unpaired data across diverse weather conditions. In addition, Xie et al. \cite{xie2025unsupervised} propose UAIR, an unsupervised all-in-one adverse weather image restoration framework that leverages contrastive learning to enhance both content preservation and category alignment without requiring paired training data.
DCD-GAN \cite{pjs} incorporates contrastive learning loss as a constraint during network training, which enhances the model performance. 
NLCL \cite{nlcl} uses a decomposition-based non-local contrastive learning strategy to compute the self-similarity of the image for restoration. 

However, once trained, existing frameworks cannot enhance the restoration capability without substantially altering their architecture or adding to the inference complexity.
To address this problem, we propose a self-collaboration unit (SC) strategy that enables the generators and restorers within the framework to achieve significant performance gains without increasing the GAN-based restoration framework's run-time complexity.

\section{Proposed Method}
\label{sec3}
In this section, we introduce an unsupervised re-boosting self-collaboration prompt GAN (RSCP$^{2}$GAN). 

\subsection{Parallel Prompt GAN for Image Restoration}
\label{parallel_prompt}
We propose a parallel prompt GAN (P$^{2}$GAN) method that ensures the model stability and effectiveness for unsupervised image restoration. 

\begin{figure}[t]
\centering
\includegraphics[width=1\linewidth]{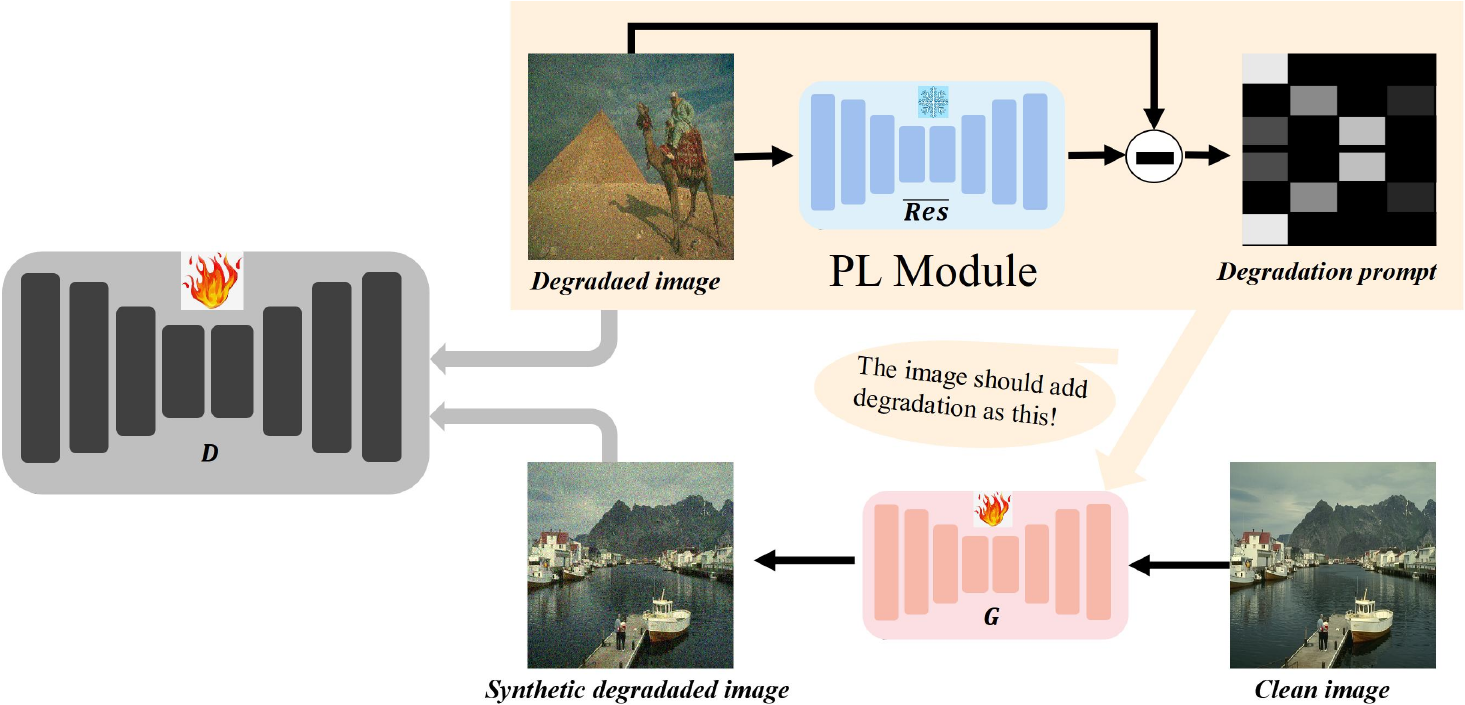}
\caption{Architecture of the prompt-guided degraded image generator. 
It learns the degradation prompt through the degradation prompt learning (PL) module and then projects the degradation prompt and clean image to the generator ($G$). 
This module reduces the burden on the $G$ when synthesizing degraded images.}
\label{shengchengqi}
\end{figure}

\begin{figure*}[t]
\includegraphics[width=1\linewidth]{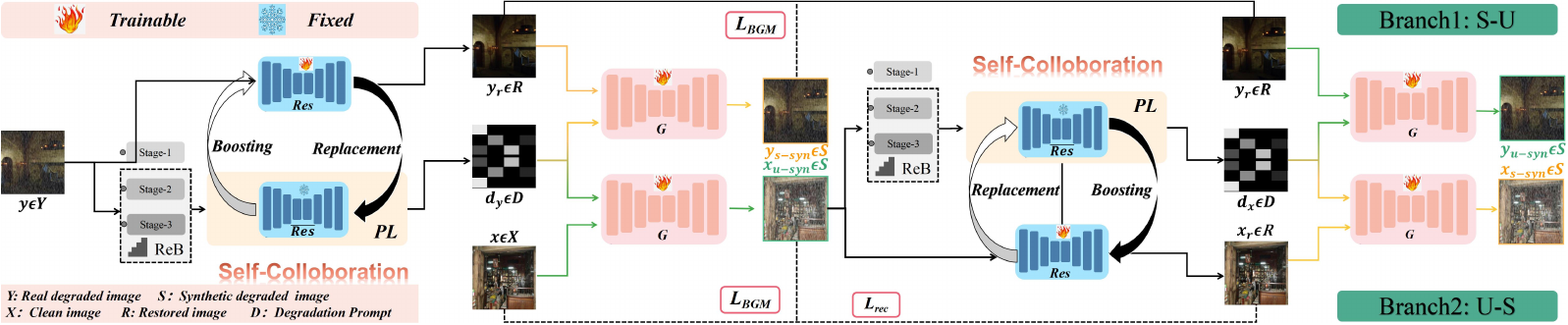}
\caption{RSCP$^{2}$GAN framework consists of two branches: Branch1: ``Self-synthesis - Unpaired-synthesis'' (Left: obtain Self-synthesis image $y_{s-syn}$ from $y_{r}$ and $d_{y}$; Right: obtain Unpaired-synthesis image $y_{u-syn}$ from $y_r$ and $d_x$) and  Branch2: ``Unpaired-synthesis - Self-synthesis" (Left: obtain Unpaired-synthesis image $x_{u-syn}$ from $x$ and $d_y$; Right: obtain Self-synthesis image $x_{s-syn}$ from $d_x$ and $x_{r}$). Each branch contains a ReB module and an SC strategy that involves a fixed restorer $\overline{Res}$ in the Prompt Learning (PL) module and a learnable restorer $Res$. This process is essentially an \textbf{R}eplacement-\textbf{B}oosting iteration, where the PL module extracts the degradation prompt from the degraded image, and the $Res$ removes the degradation from the degraded image to output a high-quality image.}
\label{kuangjia}
\end{figure*}

\subsubsection{Prompt-Guided Degraded Image Generator}
Despite the state-of-the-art performance of supervised image restoration frameworks on numerous benchmark datasets, they require a large amount of paired data, which is challenging to obtain in real-world scenarios. 
Although several GAN-based unsupervised frameworks can address this limitation, the performance is often inferior to that of supervised approaches. 
This performance gap is largely due to the domain gap between synthetic and real-world degraded images \cite{cha, jang}.
To bridge this gap, the key is to enhance the quality of synthetic images to make them as close as possible to real-world degraded images. 
If synthetic images can be generated to closely resemble real degraded images, the performance of unsupervised restoration models can approach the effectiveness of their supervised counterparts. 
Improving the quality of synthetic images is thus critical to boosting the performance of unsupervised restoration frameworks.

The generator accomplishes two tasks while synthesizing low-quality images: (1) learning the content information of real clean images, and (2) learning the degradation characteristics of real degraded images while masking their content information to avoid affecting the generation process.
As depicted in Fig. \ref{shengchengqi}, we propose a novel prompt-guided degraded image generator, which better captures degradation information. 
Since it is challenging to learn degradation information directly, instead of inputting a degraded image and a clean image into the generator \cite{hong}, we use the prompt learning (PL) module to mask the content information of the degraded images and obtain the degradation prompt. 
Specifically, a $\overline{Res}$ restores the degraded content, and the degradation prompt is obtained by subtracting the clean image from the original degraded one. 
The degradation prompt is then used to guide the generation of the synthetic degraded image with an unpaired clean image.
This approach facilitates the generator learning the image content and focusing on degradation information to synthesize degraded images, which are closer to real-world degraded ones, thereby improving the restoration performance.

\subsubsection{Parallel Prompt GAN (\texorpdfstring{P$^{2}$GAN}{P2GAN})}
\label{3.1.2}
There are two scenarios for generating pseudo-degraded images in P$^{2}$GAN:
(1) when the clean and degraded images are different, and 
(2) when the clean and degraded images are the same. 
The first scenario is a common unpaired-synthesis approach used in many unsupervised works \cite{learn, jang, hong}. 
It learns the degradation information from the degraded images and guides the generation of a pseudo-degraded image from another clean image. 
This method imposes unpaired constraints on the generator, enabling it to capture more prior information and improve the quality of pseudo-degraded images.
A robust generator should learn the real degraded properties of different inputs.
To balance the degraded content extracted from ``same image'' and ``different images'', we propose self-synthesized contents for restoration. 
These two complementary constraints improve the generator-discriminator's adversarial performance and produce pseudo-degraded images that are more consistent with the real-world degradation distribution.

As shown in Fig. \ref{kuangjia}, P$^{2}$GAN comprises two branches: branch 1 utilizes the ``self-synthesis—unpaired synthesis'' architecture, and branch 2 employs the ``unpaired synthesis—self-synthesis'' architecture. 
Specifically, branch 1 generates the self-synthesized degraded image $y_{s-syn}$ and the unpaired synthesized degraded image $y_{u-syn}$.
On the other hand, branch 2 generates the unpaired synthesized degraded image $x_{u-syn}$ and the self-synthesized degraded image $x_{s-syn}$.
These images are then fed as inputs to the discriminator, along with the real degraded image $y$. 
The ``self-synthesis'' and the ``unpaired-synthesis'' constraints are strongly complementary within and between each branch of P$^{2}$GAN.

\begin{figure}[t]
\centering
\includegraphics[width=1\linewidth]{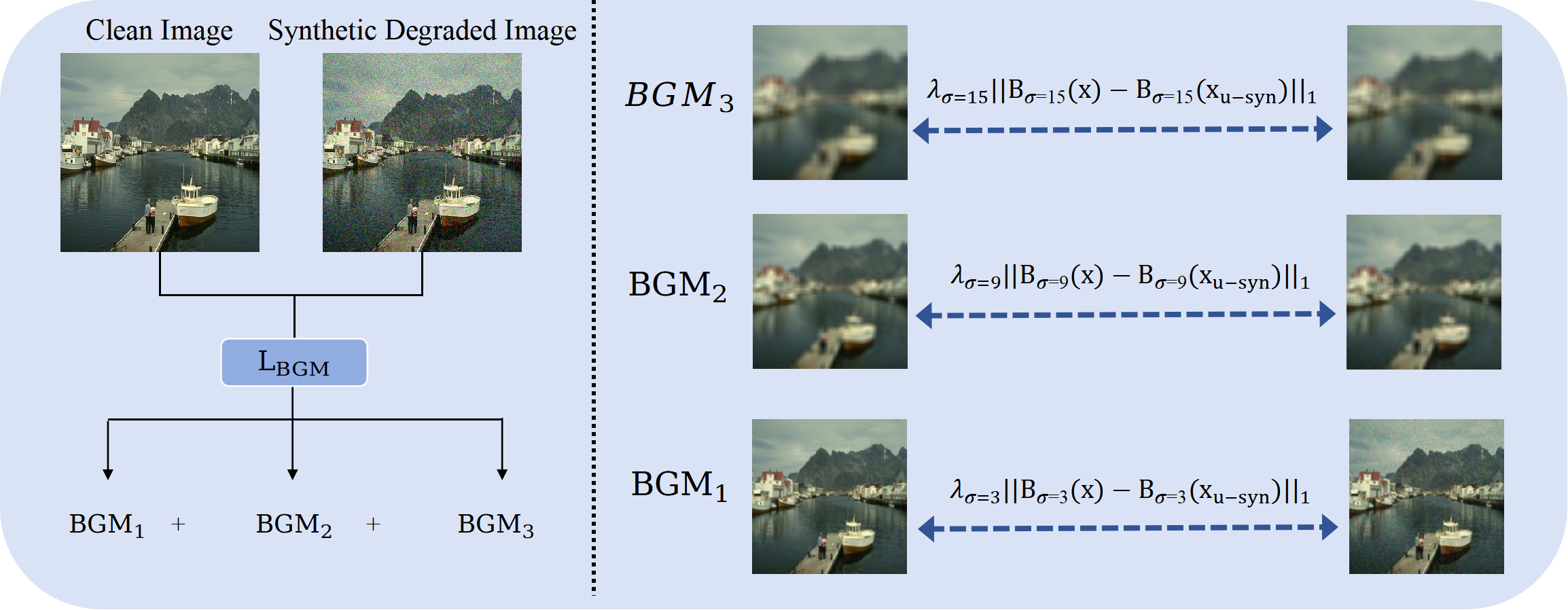}
\caption{Example of BGM loss. We use different Gaussian-Blur levels to ensure content consistency between clean images and synthetic degraded images.}
\label{bgm}
\end{figure}

\subsubsection{GAN-based Degradation Synthesis}
\label{3.1.3}
As depicted in Fig. \ref{kuangjia}, $x$ and $y$ represent the clean and degraded images. 
The generator $G$ aims to perform domain transformation by learning the image distribution in an unsupervised GAN framework. 
Simultaneously, the discriminator $D$ distinguishes whether a given degraded image is synthesized by our generator $G$ or sampled from a real degraded image dataset.
Here, $G$ and $D$ are trained adversarially to accomplish the domain transformation.
For the $x_{u-syn}$ generative process in Fig.~\ref{kuangjia}, we extract the degradation prompt $d_y$ from the real degraded image $y$ using the PL module, and input both the degradation prompt and clean image into $G$ to synthesize a synthetic degraded image $x_{u-syn}$:
\begin{equation}
x_{u-syn}= G(x, \text{PL}(y)),
\end{equation}
To prevent model degradation during training and improve the representation capability of the network, we use the adversarial loss for $L_{\text{adv2}}$:
\begin{equation}
\begin{split}
L_{\text{adv2}}=&\Vert D(y)-\mathbf{1}\Vert_{2}^{2}+\Vert D(x_{u-syn})-\mathbf{0}\Vert_{2}^{2},
\end{split} 
\end{equation}
which means for the generated image $x_{u-syn}$, its adversarial loss $L_{\text{adv2}}$ is constrained between $y$ and $x_{u-syn}$. 
The other three adversarial losses can be constructed similarly by constraining the current generated images and $y$:
\begin{equation}
\begin{aligned}
L_{\text{adv1}} &= \Vert D(y)-\mathbf{1} \Vert_{2}^{2} + \Vert D(y_{s\text{-}syn}) - \mathbf{0} \Vert_{2}^{2} \\
L_{\text{adv3}} &= \Vert D(y)-\mathbf{1} \Vert_{2}^{2} + \Vert D(y_{u\text{-}syn}) - \mathbf{0} \Vert_{2}^{2} \\
L_{\text{adv4}} &= \Vert D(y)-\mathbf{1} \Vert_{2}^{2} + \Vert D(x_{s\text{-}syn}) - \mathbf{0} \Vert_{2}^{2}
\end{aligned}
\end{equation}
The overall loss for the GAN model is:
\begin{equation}
L_{\text{GAN}} = L_{\text{adv1}}+L_{\text{adv2}}+L_{\text{adv3}}+L_{\text{adv4}}.
\end{equation}

Similar to \cite{bgm}, we apply a background guidance module (BGM) to provide additional supervision. 
The BGM maintains the consistency of the background between the synthetic degraded image and the clean image, constraining their low-frequency contents to be similar.
We illustrate this approach using $L_{\text{BGM}}$ in branch 2. 
Low-frequency contents are extracted by using several low-pass filters and constrained to be close to each other through the L1 loss:
\begin{equation}
\begin{split}
L_{\text{BGM}}=\sum_{\sigma=3,9,15}\lambda_{\sigma}\Vert B_{\sigma}(x)-B_{\sigma}(x_{u-syn})\Vert_{1},
\end{split} 
\end{equation}
where $B_{\sigma}$ denotes the Gaussian filter operator with blurring kernel size $\sigma$, and $\lambda_{\sigma}$ denotes the weight for the level $\sigma$.
An example of the BGM loss is shown in Fig. \ref{bgm}. We set $\sigma$-s to 3, 9, and 15, and $\lambda$-s to 0.01, 0.1, and 1, respectively.

In the image restoration framework, we utilize pseudo-paired samples denoted by $x^{i}$ and $x_{rec}^{i}$. 
The restorer is trained by optimizing the following loss functions:
\begin{equation}
\begin{split}
\label{ssim}
L_{\text{Res}}=&\frac{1}{2m}\sum_{i=1}^m\Bigg[\Vert x_{rec}^{i}-x^{i}\Vert_{1}
+ \lambda_{\text{SSIM}}L_{\text{SSIM}}(\mathbf{x}_{rec}^{i}, x^{i})\Bigg],
\end{split} 
\end{equation}
where $m$ denotes the total number of the sample pairs, $x_{rec}^{i}$ is the clean image estimated by the restorer, $L_{\text{SSIM}}$ represents the structural information used by SSIM loss to constrain the image, and $\lambda_{\text{SSIM}}$ is the weight for $L_{\text{SSIM}}$.
The total loss function is:
\begin{equation}
L = \min_{G}\max_{D} \left[L_{\text{GAN}}+\lambda_{\text{BGM}}L_{\text{BGM}} + L_{\text{Res}}\right],
\end{equation}
where $\lambda_{\text{BGM}}$ is the weight of background consistency loss.

\subsection{Proposed SC Based \texorpdfstring{P$^{2}$GAN}{P2GAN}}
\label{sec3_3}
As P$^{2}$GAN introduced in Section \ref{parallel_prompt} is formulated within the conventional GAN-based unsupervised framework, it is challenging to achieve further performance gains without significantly modifying the architecture or increasing inference complexity. 
To address these issues, we propose the SC-based P$^{2}$GAN (SCP$^{2}$GAN) model. 

\begin{figure}[t]
\centering
\includegraphics[width=1\linewidth]{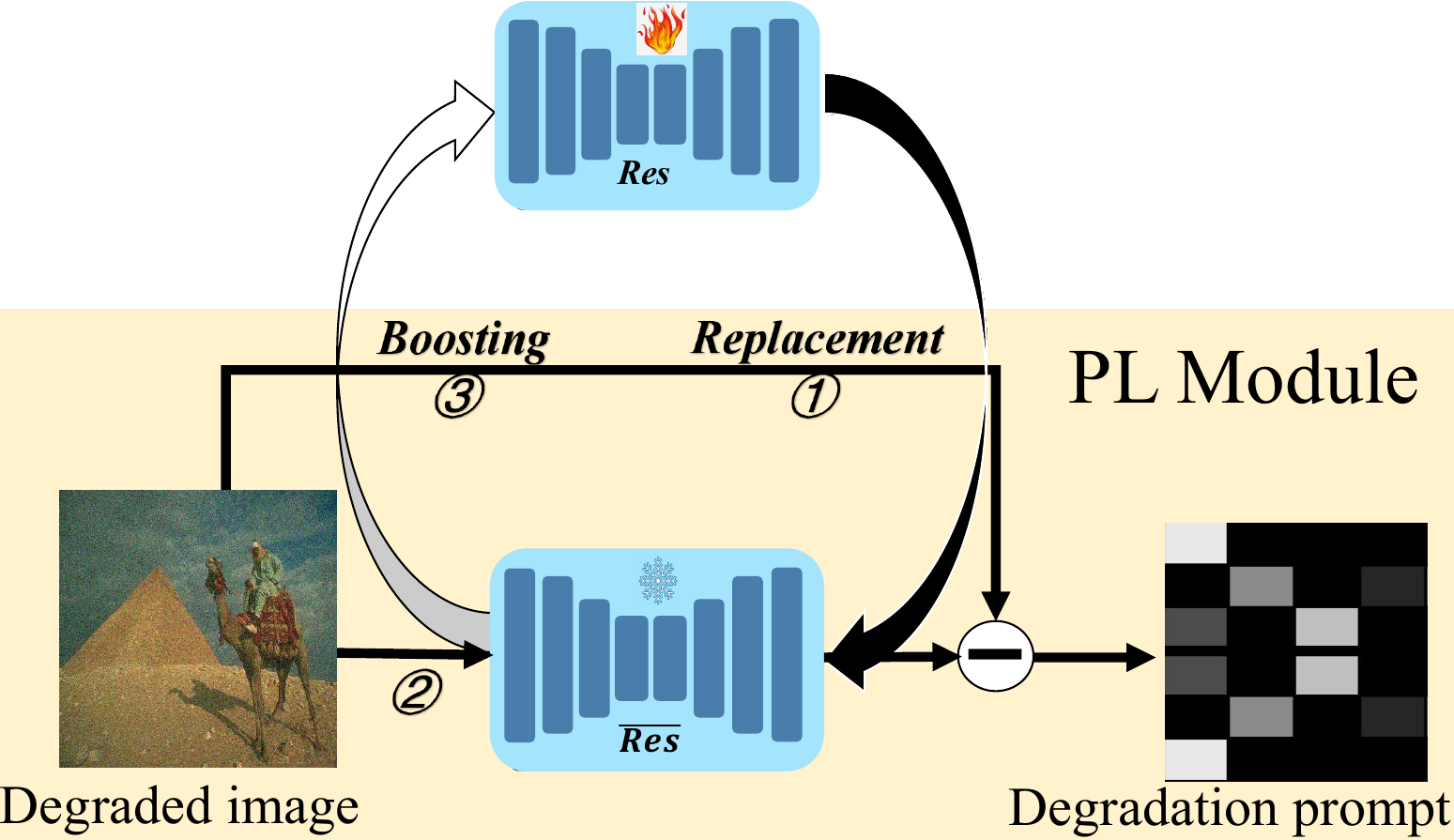}
\caption{Illustration of the SC strategy. The $k$ represents the number of SC iterations. In each iteration, the previously trained $Res$ is fixed and used as the new $\overline{Res}$ to start the next iteration of the Reb-SC. In the PL module, the $\overline{Res}$ is a learnable convolutional block when $k$=0.
When $k$\textgreater0, the iterative collaboration of \textbf{$\overline{Res}$} in the PL module and the restorer $Res$ is conducted. 
\textcircled{\scriptsize{1}}: the current restorer $Res$ replaces the previous weaker $\overline{Res}$, which enhances the performance of the degradation prompt learning (PL) module.
\textcircled{\scriptsize{2}}: the updated PL can generate better clean-degraded image pairs used to train the $Res$.
\textcircled{\scriptsize{3}}: the $Res$ is trained using the updated clean-degraded image pairs, further boosting its performance.}
\label{Fig4}
\end{figure}

\subsubsection{Self-Collaboration Strategy}
\label{3.2.1}
The proposed Self-collaboration (SC) strategy enables a restorer trained in a conventional unsupervised framework to self-correct and improve its performance without requiring modifications to its structure or increased complexity in the inference phase.
The SC strategy, illustrated in Fig. \ref{Fig4}, comprises a prompt learning (PL) module and a restorer ($Res$). 
As described above, the PL module extracts degradation prompts from a real degraded image and guides the generator to produce high-quality pseudo-degraded images. 
To train the P$^{2}$GAN, we initially use a simple and learnable linear convolutional layer as the $\overline{Res}$ in the PL module. 
Then, the $Res$ is iteratively replaced and boosted. 
During each iteration, the current more powerful $Res$ replaces the previous weaker $\overline{Res}$ in the PL module. 
This leads to a more effective $\overline{Res}$ to extract more accurate degradation prompts. 
That is, it generates more realistic synthetic degraded/clean image pairs and iteratively improves the performance of the updated $Res$ with higher-quality synthetic samples. 
We observe a significant improvement in $Res$'s performance using the SC strategy compared to the original one without SC.
During the SC stage, we set the 
\begin{equation}
R_{fake1} = \overline{Res}(x_{u-syn}), 
R_{fake2} = \overline{Res}(y), \\
\end{equation}
and the loss functions of $G$ and $D$ are the same as before. 
The loss function of $Res$ is defined as:
\begin{equation}
\begin{split}
\label{SC}
L_{\text{Res-SC}}= &L_{\text{Res}} + \Vert x_{r}-R_{fake1}\Vert_{1}\\
&+ \Vert y_{r}-R_{fake2}\Vert_{1} \\
&+\lambda_{\text{SSIM}}L_{\text{SSIM}}(x_{r}, R_{fake1}) \\
&+\lambda_{\text{SSIM}}L_{\text{SSIM}}(y_{r}, R_{fake2}).
\end{split} 
\end{equation}

Using feedback from one part of a framework to improve other parts is called positive feedback.
It encourages our SC (self-collaboration) strategy of utilizing feedback from one part to guide the improvement of other parts. 
Subsequently, the improved parts can, in turn, guide the initial parts. 
This iterative process of positive feedback is referred to as self-collaboration. 
%
We show this approach can facilitate numerous low-level vision tasks.

\begin{figure}[t]
\centering
\includegraphics[width=1\linewidth]{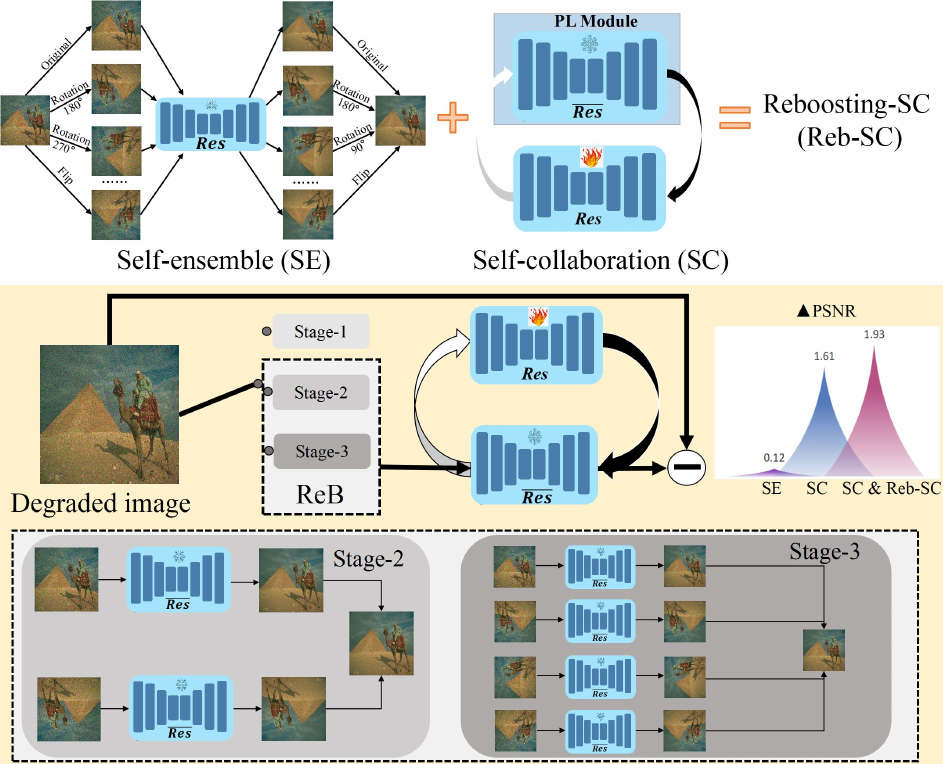}
\caption{Illustration of the Reb-SC strategy. The self-ensemble (SE) can improve the model's performance during testing, but it increases the network's computational complexity. In contrast, our self-collaboration (SC) is used only in the training phase, does not increase inference time, and can significantly enhance the restorer's performance. Based on this, we combine the characteristics of the SE and SC to propose Reb-SC. In Stage 1, we use the original SC; in Stage 2 and Stage 3, we use Reb-SC. We input augmentation of low-quality images into the fixed PL module to enhance the performance of the $\overline{Res}$ in PL, thereby training a better restorer $Res$. The SE can bring a 0.12 dB improvement to the $Res$, while our SC can provide a 1.61 dB improvement. However, the Reb-SC combines both characteristics and can further improve the performance of the restorer, achieving a 1.93 dB improvement.}
\label{rebsc}
\end{figure}

\subsubsection{Re-Boosting SC Strategy}
The typical self-ensemble (SE) strategy applies random flip and rotation to input images and averages the resulting outputs to achieve better performance during testing \cite{aind1, 74}. 
However, this approach increases inference times and provides only limited improvements. 
In contrast, the proposed SC strategy avoids additional test computation while delivering significant improvements with only minor modifications to the training phase. 
In this work, we propose a Re-boosting SC (Reb-SC) module that combines the SC and SE strategies.
Specifically, as shown in Fig. \ref{rebsc}, the Reb-SC strategy is applied at the end of the SC process: the original input to a fixed PL undergoes self-ensemble with multiple inputs. 
The outputs are averaged, leading to improved performance of the PL and further enhancement of the $Res$. 
The process is:
\begin{align}
&x_{u-syn-(1,2,...,k)} = Aug(x_{u-syn}), \\
&R_{fake1-(1,2,...,k)} = \overline{Res}(x_{u-syn-(1,2,...,k)}), \\
&y_{1,2,...,k} = Aug(y), \\
&R_{fake2-(1,2,...,k)} = \overline{Res}(y_{1,2,...,k}), \\
&R_{fake1} = \dfrac{1}{k}\sum_{i=1}^k(R_{fake1-(1,2,...,k)}), \\
&R_{fake2} = \dfrac{1}{k}\sum_{i=1}^k(R_{fake2-(1,2,...,k)}), 
\end{align}
where the $k$ is the number of augmentation images, the loss function of the $Res$ is the same as the SC strategy before.

\begin{figure*}[t]
\centering
\includegraphics[width=1\linewidth]{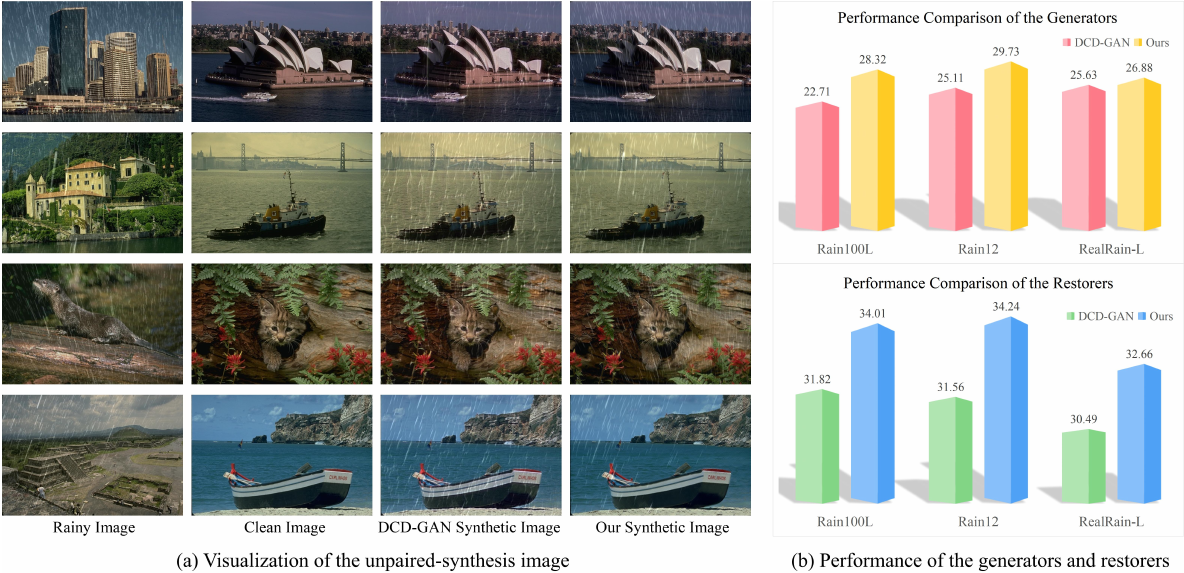}
\caption{(a) Visualization of the unpaired synthetic degraded image by DCD-GAN \cite{pjs} and our RSCP$^{2}$GAN. (b) Quantitative comparison between DCD-GAN and our framework for the generator and the restorer.}
\label{shengcheng}
\end{figure*}

\begin{algorithm}[t]
\small
\caption{The training process of our RSCP$^{2}$GAN.}
\label{alg}
\begin{algorithmic}[1]

\Require{$Res$: the restorer; $Gau$: the Gaussian filter;
           $\overline{Res}$}: the restorer in the PL module; $D$: degraded images; $R$: restored images; 
           $s_{1,2,3}$: the last epoch of stage 1,2,3; $s$: current numbers of epoch

\vspace{-5pt}
$\quad\quad$\tikzmk{A} \Comment{The basic stage without SC.}

\textbf{if} $s < s_{1}$ \textbf{then} $\overline{Res}$ = $Gau$

    \textbf{for} epoch in 0 to $s_{1}$ epochs \textbf{do:}

        ~~~~~$R$ $\leftarrow$ $Res$($D$)    

        ~~~~~Optimizer($Res$, $R$)

\vspace{-5pt}\hspace{-0.2cm}\tikzmk{B}\boxit{pink}

\vspace{-10pt}
$\quad\quad$\tikzmk{A} \Comment{The SC stage.}

\textbf{else if} $s_{1} < s < s_{2}$ \textbf{then} $\overline{Res}$ = $Res$

\textbf{for} epoch in 0 to $s_{2} - s_{1}$ epochs \textbf{do:}  

    ~~~~~$R_{fake}$ $\leftarrow$ $\overline{Res}$($D$)

    ~~~~~$R$ $\leftarrow$ $Res$($D$)

    ~~~~~Optimizer($Res$, $R$, $R_{fake}$)
    
\vspace{-5pt}\hspace{-0.2cm}\tikzmk{B}\boxit{blue2}

\vspace{-10pt}
$\quad\quad$\tikzmk{A} \Comment{The Reb-SC stage.}

\textbf{else if} $s_{2} < s < s_{3}$ \textbf{then} $\overline{Res}$ = $Res$

\textbf{for} epoch in 0 to $s_{3} - s_{2}$ epochs \textbf{do:}

    ~~~~~\textbf{for} k in 0...max folds \textbf{do}

        ~~~~~~~~~~$D_{1,2,...,k}$ $\leftarrow$ Aug($D$)

        ~~~~~~~~~~$R_{fake-(1,2,...,k)}$ $\leftarrow$ $\overline{Res}$($D_{1,2,...,k}$)
        
    ~~~~~$R_{fake}$ $\leftarrow$ $R_{fake-(1,2,...,k)}$

    ~~~~~$R$ $\leftarrow$ $Res$($D$)

    ~~~~~Optimizer($Res$, $R$, $R_{fake}$) 
    
\vspace{-5pt}\hspace{-0.2cm}\tikzmk{B}\boxit{green} 

\end{algorithmic}
\end{algorithm}

\subsubsection{Deep Analysis of \texorpdfstring{RSCP$^{2}$GAN}{RSCP2GAN}}

We propose RSCP$^{2}$GAN by 
integrating the SC and Reb-SC strategies with our P$^{2}$GAN. 
The training process of our RSCP$^{2}$GAN is detailed in Algorithm \ref{alg}. 
In this model, $Res$ represents the trainable restorer, $\overline{Res}$ is the fixed restorer within the PL module,  and $Gau$ denotes the Gaussian filter. 
The symbols $D$, $R$, $s_{1,2,3}$, and $s$ refer to degraded images, restored images, and the last epochs of stages 1, 2, 3, and the current epoch, respectively.
Initially, we train P$^{2}$GAN through the first stage until the restorer converges.

\begin{table*}[tp]
\begin{center} 
\fontsize{9}{9}\selectfont
\renewcommand{\arraystretch}{1.6}
\caption{Denoising results of several competitive methods on SIDD Validation, SIDD Benchmark, and DND Benchmark.  
Additionally, $*$ denotes that the approach is trained on the DND benchmark directly, and the results without $*$ means the methods trained on the SIDD datasets.} 
\label{denoising_quantitative}
  \centering
\setlength{\tabcolsep}{2.2mm}{
\begin{tabular}{lllcccccc}
\toprule
&\multirow{2}[3]{*}{Methods} & GAN-based &\multicolumn{2}{c}{SIDD Validation} 
& \multicolumn{2}{c}{SIDD Benchmark} & \multicolumn{2}{c}{DND Benchmark} \\
\cmidrule{4-9} & & /Publication & PSNR$\uparrow$ & SSIM$\uparrow$ & PSNR$\uparrow$ & SSIM$\uparrow$ & PSNR$\uparrow$ & SSIM$\uparrow$\\
\midrule
\multirow{2}{*}{Non-learning} & BM3D \cite{5} & No/TIP 2007 & 31.75 & 0.7061 & 25.65 & 0.6850 & 34.51 & 0.8510 \\
& WNNM \cite{12}& No/CVPR 2014 & $-$ & $-$ & 25.78 & 0.8090 & 34.67 & 0.8650 \\
\midrule
\multirow{7}{*}{\shortstack{Real pairs\\(Supervised)}}
& TNRD \cite{51} & No/TPAMI 2016 & 26.99 & 0.7440& 24.73 & 0.6430 & 33.65 & 0.8310 \\
& DnCNN \cite{17} & No/TIP 2017 & 26.20 & 0.4414 & 28.46 & 0.7840 & 32.43 & 0.7900 \\
& RIDNet \cite{ridnet}& No/CVPR 2019 & 38.76 & 0.9132 & 37.87 & 0.9430 & 39.25 & 0.9530 \\
& AINDNet \cite{aind2} & No/CVPR 2020 & 38.96 & 0.9123 & 38.84 & 0.9510 & 39.34 & 0.9520 \\
& DeamNet \cite{aind1} & No/CVPR 2021 & 39.40 & 0.9169 & 39.35 & 0.9550 & 39.63 & 0.9531 \\
& ScaoedNet \cite{74} & No/NeurIPS 2022 & 39.52 & 0.9187 & 39.48 & 0.9570 & 40.17 & 0.9597 \\
& Restormer \cite{74} & No/CVPR 2022 &  39.93 & 0.960 & 40.02 & 0.960 & 40.03 & 0.956 \\
\midrule
\multirow{5}{*}{\shortstack{Synthetic pairs \\ (Two Stages)}} & DnCNN \cite{17} & No/TIP 2017 & $-$ & $-$ & 23.66 & 0.5830 & 32.43 & 0.7900 \\
& CBDNet \cite{25} & No/CVPR 2019 & 30.83 & 0.7541 & 33.28 & 0.8680 & 38.06 & 0.9420 \\
& PD+ \cite{pd} & No/AAAI 2020 & 34.03 & 0.8810 & 34.00 & 0.8980 & 38.40 & 0.9450 \\
& C2N$+$DnCNN \cite{jang}& Yes/ICCV 2021& $-$ & $-$ & 33.76 & 0.9010 & 36.08 & 0.9030 \\
& C2N$+$DIDN \cite{jang}& Yes/ICCV 2021 & $-$ & $-$ & 35.02 & 0.9320 & 36.12 & 0.8820 \\
\midrule
\multirow{12}{*}{Unsupervised}& N2V \cite{n2v} & No/CVPR 2019 & 29.35 & 0.6510 & 27.68 & 0.6680 & $-$ & $-$ \\
& GCBD \cite{chen} & Yes/CVPR 2018 & $-$ & $-$ & $-$ & $-$ & 35.58 & 0.9220 \\
& UIDNet \cite {hong} & Yes/AAAI 2020 & $-$ & $-$ & 32.48 & 0.8970 & $-$ & $-$ \\
& R2R \cite{r2r} & No/CVPR 2021 & 35.04 & 0.8440 & 34.78 & 0.8980 & 36.20 & 0.9250 \\
& CVF-SID ($S^{2}$) \cite{cvf}& No/CVPR 2022 & $-$ & $-$ & 34.71 & 0.9170 & 36.50 & 0.9240 \\
& AP-BSN+$R^{3}$ \cite{apbsn} & No/CVPR 2022 & 35.76 & $-$ & 35.97 & 0.9250 & 38.09 & 0.9371 \\
& LG-BPN+$R^{3}$ \cite{lgbpn} & No/CVPR 2023 & 37.31 & 0.8860 & 37.28 & 0.9360 & 38.02 & 0.9373 \\
& BNN-LAN \cite{bnnlan} & No/CVPR 2023 & 37.39 & 0.8830 & 37.41 & 0.9340 & 38.18 & 0.9386 \\
& SDAP (E) \cite{pyz} & No/ICCV 2023 & 37.30 & 0.8937 & 37.24 & 0.9360 & 37.86 & 0.9366 \\
& SCPGabNet \cite{scpgabnet} & Yes/ICCV 2023 & 36.53 & 0.8860 & 36.53 & 0.9250 & 38.11 & 0.9393 \\
& PUCA \cite{puca} & No/NeurIPS 2024 & 37.49 & 0.8800 & 37.54 & 0.9360 & 38.83* & 0.9420* \\
& Complementary-BSN \cite{tcsvt_denoising} & No/TCSVT 2024 & 37.51 & 0.8850 & 37.43 & 0.9360 & 38.24 & 0.9400 \\
& RSCP$^{2}$GAN (ours) & Yes/$-$ & 37.83 & 0.9070 & 37.69 & 0.9450 & 38.37 & 0.9421 \\
\bottomrule
\label{table0}
\end{tabular}}%
\end{center}
\end{table*}%

In the SC stage, at the beginning of each iteration, the new PL more accurately captures the degradation prompt in the degraded image by replacing $\overline{Res}$ in the PL module with an improved $Res$. 
This reduces the influence of image content on the synthetic degraded image generation process. 
As illustrated in Fig. \ref{kuangjia}, with more precise degradation prompt extraction from the degraded image $y$, our $Res$ achieves better results in both self-synthesis in branch 1 and unpaired synthesis in branch 2, leading to higher-quality synthetic degraded images. 
Similarly, a more accurate degradation prompt extracted from the synthetic degraded image $x_{u-syn}$ improves unpaired synthesis in branch 1 and self-synthesis in branch 2, thus enhancing complementary constraints between the two branches and improving the interconnectedness of the network modules. 
Consequently, our SC strategy establishes a self-boosting framework that enhances $Res$ training and performance. 
The implementation of the SC strategy involves several steps:
After the original P$^{2}$GAN framework has converged, replace $\overline{Res}$ in the PL module with the latest $Res$ and fix its parameters to generate better pseudo-degraded images.
Retrain $G$, $D$, and $Res$ until convergence is achieved. Repeat this process until the performance of $Res$ no longer improves. 
In the Reb-SC stage, we use the Reb-SC strategy to further enhance the performance of the PL module and $Res$. 
Specifically, we augment synthesized low-quality images and input them into the fixed $\overline{Res}$ within the PL module to improve its performance, thereby training a better-performing $Res$.

We validate the effectiveness of the pseudo-degraded images generated by RSCP$^{2}$GAN.
As shown in Fig. \ref{shengcheng}(a), we compare the real degraded image, the unpaired clean image, degraded images generated by the current state-of-the-art algorithm DCD-GAN \cite{pjs}, and degraded images generated by our method. 
The rain streaks produced by DCD-GAN \cite{pjs} are curved and do not align with the real degraded image, which is a core reason for the unsatisfactory performance of the restorer. 
To quantitatively assess the impact of the generated images, we evaluate cases where the degraded image and the clean image originate from the same image content.
As shown in the upper part of Fig. \ref{shengcheng}(b), our generator outperforms DCD-GAN across multiple datasets. 
Additionally, the lower part of Fig. \ref{shengcheng}(b) demonstrates that our restorer surpasses DCD-GAN, further indicating that improving the generator performance is an effective way to enhance the restorer performance.

\begin{table*}[t]
\begin{center} 
\fontsize{8.5}{8.5}\selectfont
\caption{Qualitative Comparison on the PolyU Dataset.} 
\label{denoising_polyu}
\renewcommand{\arraystretch}{1.6}
\setlength{\tabcolsep}{1.0mm}{
\begin{tabular}{ccccccccc}
\toprule
Method & CVF-SID \cite{cvf} & AP-BSN+$R^{3}$ \cite{apbsn} & BNN-LAN \cite{bnnlan} & LG-BPN+$R^{3}$ \cite{lgbpn} & SCPGabNet \cite{scpgabnet} & SDAP (E) \cite{pyz} & RSCP$^{2}$GAN (ours) \\
\midrule
PSNR/SSIM & 33.00/0.9101 & 36.88/0.9496 & 37.13/0.9541 & 36.25/0.9473 & 37.14/0.9534 & 37.21/0.9537 & 37.61/0.9549 \\
\toprule
\label{polyu}
\end{tabular}}%
\end{center}
\end{table*}%

\begin{figure*}[t]
\centering
\includegraphics[width=1\linewidth]{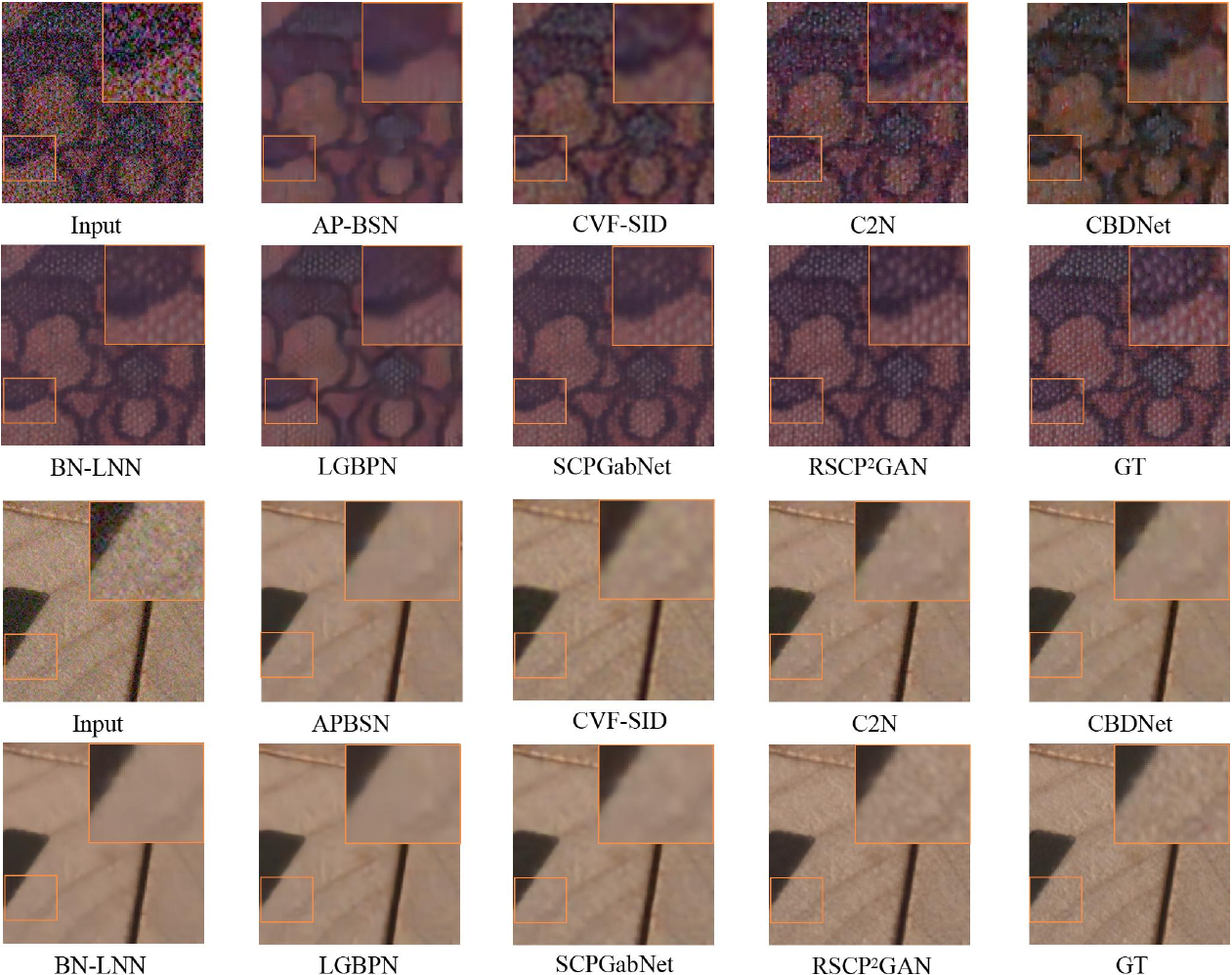}
\caption{
Visual comparison of our method against other competing methods on the SSID \cite{sidd} Validation.}
\label{quzaoduibi}
\end{figure*}

\begin{figure*}[t]
\centering
\includegraphics[width=1\linewidth]{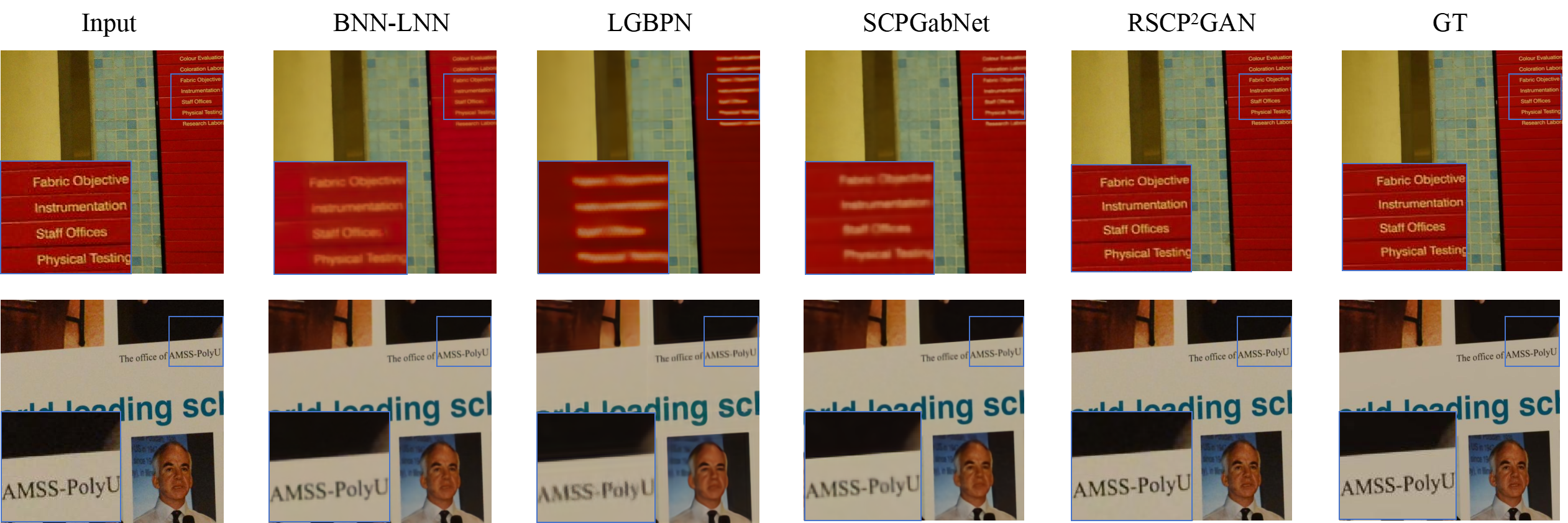}
\caption{
Visual comparison of our method against other competing methods on the PolyU \cite{polyu}.}
\label{quzaoduibi_polyu}
\end{figure*}

\subsubsection{Analysis of the Proposed SC and Reb-SC}
\label{analysis_rebsc}
Traditional data augmentation is applied during the testing phase of a trained model, where data is augmented, followed by multi-branch restoration and averaging. This significantly increases both the testing time and memory usage, which is disadvantageous when applying restoration networks in the real world. In contrast, our re-boosting applies data augmentation during the training phase, further enhancing the proposed SC. Like the original SC, it is applied during the training phase, making only minor changes to the framework, without affecting testing time or memory usage in the reference phase, and having no impact on the application of restoration networks.

\textbf{Training:} Taking the original framework without self-collaboration as a reference, suppose it requires training for $T_0$ epochs, with a per-epoch complexity of $P_0$, which contains the complexity from the generator, discriminator, and restorer. The total complexity of training the original framework $P_{Ori_{Train}}
= T_0 \times P_0$.

After applying SC, the training runs for additional $T_s$ epochs ($T_s \ll T_0$), with each epoch having a complexity of $P_s$. Since the parameter $\overline{Res}$ in the SC is fixed, and is larger than the initial simple Gaussian filter in the basic stage without SC of the P$^{2}$GAN (See Algorithm \ref{alg}), while the complexity of the generator and discriminator is not changed, it only adds about (slightly lower) $\overline{Res}$'s complexity. Therefore, we get $P_s$ higher than $P_0$, but much less than $2 \times P_0$, i.e., $P_0 < P_s < 2 \times P_0$. Thus, the complexity in the SC phase is $P_{SC_{Train}}=T_{s}\times P_{s}$, where $T_{s}\times P_{0}<P_{SC_{Train}}<2\times T_{s} \times P_{0}$. As shown in Fig. \ref{rebsc}, for the Reb-SC, we have two or four input images (the degraded images with different augmentation) to the $\overline{Res}$, and run only one epoch in each stage. Thus, the complexity $P'_s$ of the framework in Reb-SC is less than (4 + 2) $\times P_s$ = 6 $\times P_s$ and larger than $2 \times P_{0}$. This introduces additional complexity for the Reb-SC stage, denoted as $P_{REB_{Train}}$, such that: $2 \times P_{0} < P_{REB_{Train}} < 6 \times P_{s}<2\times6\times P_{0}=12 \times P_{0}$. Thus, the overall training complexity of the framework becomes $P_{Train} = P_{Ori_{Train}} +  P_{SC_{Train}}+P_{REB_{Train}}$, where $T_{0} \times P_{0} + T_{s} + 2\times P_{0} = (T_{0}+T_{s}+2)\times P_{0}<P_{Train}<T_{0}\times P_{0}+2\times T_{s}\times P_{0} + 12 \times P_{0}=(T_{0}+2\times T_{s}+12)\times P_{0}$. Since ($2 \times T_s + 12$) is relatively small compared to  $T_0$, the added complexity $(2 \times T_s + 12) \times P_{0}$ is also smaller than the original complexity $T_{0} \times P_{0}$, resulting in a slight increase in training complexity.

\textbf{Inference:} The same restoration module is used as in the non-Reb-SC version, so the inference complexity is simply that of a single \textit{denoising} pass, denoted as $P_{SC\&REB_{Inference}} = P_{Inference}$. In traditional self-ensemble augmentation strategies, $N$ augmented variants are typically averaged during inference, leading to a complexity of $P_{Aug_{Inference}} = N \times P_{Inference}$
$N \times P_0$ complexity. When $N = 8$, such traditional self-ensemble augmentation significantly increases the inference complexity to $8\times P_{0}$.

\textbf{Performance Improvement:} As we presented in Fig. \ref{shengcheng}, under the CycleGAN framework, the performance of the restorer depends on the quality of the data generated by the generator — the better the generated data, the better the trained restorer. There is a strong positive correlation between them. The quality of the generator, in turn, depends on the accuracy of the degradation hint information in the input. Taking denoising as an example, this refers to the accuracy of noise estimation. In our framework, we compute the noise information by subtracting the clean image (restored by the current restorer) from the real noisy image. Thus, the better the current restoration, the more accurate the extracted noise information becomes. More accurate noise information leads to more precise pseudo-noisy images, which in turn further improve the performance of the restorer. Consequently, the denoising performance will gradually improve with our method. The performance improvement analyses of other restoration tasks using our framework are similar.

Specifically, let $Q_t$ be the current restoration performance of the restorer in our framework, $Q_{t-1}$ and $Q_{t+1}$ denote the previous and next SC/Reb-SC iteration performance of the restorer. Let $P_t$ be the current quality of the input prompt to the generator, which is obtained by the degraded image subtract the previous restored image by the restorer with restoration performance $Q_{t-1}$. Let $G_t$ be the current performance of the generator. Firstly, we have $Q_0$, $P_0$, and $G_0$, where $Q_0$ is the performance of a simple filter (as illustrated in Sec. \ref{3.2.1} and Algorithm \ref{alg}). When applying SC, we use a better restorer with $Q_1$ ($Q_{1}$ refers to the performance of the restoration network NAFNet trained under the complete unsupervised framework without SC or Reb-SC. Thanks to NAFNet's advanced design and its deep exploitation of data characteristics, it achieves superior restoration performance.) to replace the previous filter with $Q_0$ ($Q_{0}$ represents the performance of a simple Gaussian filter with a basic convolutional structure and limited fitting capacity.) in the framework. Then we get a better prompt $P_1$, which can further train a better generator with $G_1$. After this SC iteration, we get a better restorer with $Q_2$. Following the same process, we then get an improved prompt $P_2$ and generator with $G_2$. 
Similarly, we can conclude that if the restorer with $Q_{T_{SC}}$ is better than $Q_{T_{SC}-1}$ ($Q_{T_{SC}-1}$ is used in the input of the
generator with $G_{T_{SC}-1}$, and $T_{SC}$ is the current iteration number of the SC phase), replacing the restorer from $Q_{T_{SC}-1}$ to $Q_{T_{SC}}$ can generate better prompt $P_{T_{SC}}$ (i.e., $P_{T_{SC}} > P_{T_{SC}-1}$). A better $P_{T_{SC}}$ leads to a stronger generator with $G_{T_{SC}}$ (i.e., $G_{T_{SC}} > G_{T_{SC}-1}$), thus improving the restoration performance. By continuously iterating in this manner, the performance improves until convergence.

At the Reb-SC stage, we have two or four input images (the degraded images with different augmentations) to the $\overline{Res}$ and average the restored results to get a better restored image. Let $T_{Reb-SC}$ be the current iteration number of the Reb-SC phase. Then, the input of the generator with $G_{T_{Reb-SC}-2}$ will be improved since multiple restored results generated by the restorer with $Q_{T_{Reb-SC}-1}$ are averaged, leading to a better $P_{T_{Reb-SC}-1}$. The better $P_{T_{Reb-SC}-1}$ will enhance the generator further with $G_{T_{Reb-SC}-1}$, and gets a better restorer $Q_{T_{Reb-SC}}$. After this process, we get the final restorer with the best performance.

\section{Experiments}
We first describe the datasets we utilized and present the implementation details. 
Next, we provide the image denoising and deraining analysis with the existing state-of-the-art unsupervised approaches qualitatively and quantitatively. 
We conduct ablation studies to validate the effectiveness of the proposed methods and modules.

\subsection{Datasets}
\label{sec4_1}
\noindent\textbf{Denoising Task.} We conduct experiments on widely used real-world image denoising datasets: SIDD \cite{sidd}, DND \cite{dnd}, and PolyU \cite{polyu}. 
The SIDD Medium training set consists of 320 pairs of noisy and corresponding clean images captured by multiple smartphones. 
The SIDD validation and benchmark sets each contains 1280 color images of size 256 $\times$ 256. 
There are 50 high-resolution noisy images and 1000 sub-images of size 512 $\times$ 512 in the DND dataset.
The PolyU dataset contains 40 high-resolution noisy-clean image pairs for training and 100 images of size 512 $\times$ 512 for testing. 
We train our model on the SIDD training set and test it on the SIDD Validation, SIDD Benchmark, and DND Benchmark. 
Specifically, we divide the SIDD Medium training set equally into noisy and clean image parts. 
Then, we use 160 clean images from the first part and 160 noisy images from the second part to construct an unpaired dataset for training. 
Additionally, we train our approach on the PolyU training dataset and test on its testing set, following a similar processing method as for the SIDD dataset.

\noindent\textbf{Deraining Task.} We train and test our model on commonly used deraining datasets: Rain100L \cite{rain100}, RealRainL \cite{realrain}, and Rain12 \cite{rain12}. 
The Rain100L dataset has 200 synthetic image pairs for training and 100 image pairs for testing. 
The RealRainL set \cite{realrain} consists of 784 real-world image pairs for training and 224 image pairs for testing. 
The Rain12 dataset contains 12 pairs of rainy and clean images. 
Following the recent work \cite{deraincyclegan}, we test on the Rain12 dataset \cite{rain12} using models trained on the images from the Rain100L dataset.

\noindent\textbf{Desnowing Task.} We train and test our model on commonly used desnowing datasets: CSD \cite{csd}, and Snow100K \cite{snow100k}. The CSD contains 8000 image pairs for training and 2000 image pairs for testing. For the Snow100K, we use the presented 50000 image pairs for training and randomly select 2000 samples for testing. To ensure fair comparison under an unsupervised training setting, we apply the same training protocol as used in our denoising and deraining experiments.

\subsection{Implementation Details}
To optimize the proposed network, we use the Adam optimizer with $\beta_1$=0.9, $\beta_2$=0.999, and an initial learning rate of $2 \times 10^{-4}$. 
The proposed models are implemented using PyTorch and trained on two Nvidia GeForce RTX 3090 GPUs. 
For the denoising task, the batch size and patch size are set to 6 and 112, respectively. 
We set $\lambda_{\text{BGM}}$ in Eq. \ref{bgm} to 6 and $\lambda_{\text{SSIM}}$ in Eq.~\ref{ssim} to 1. 
We use a ResNet with 6 residual blocks as the generator, a PatchGAN \cite{patch} as the discriminator, and a Restormer \cite{restormer} as the restorer. 
For the deraining and desnowing tasks, following prior work \cite{pjs}, we set the batch size to 1 and the patch size to 256. The NAFNet \cite{naf} as our restorer.

\subsection{Image Denoising}\label{sec4_2}

We evaluate our method on real-world noisy images from the SIDD Validation \cite{sidd}, SIDD Benchmark \cite{sidd}, and DND Benchmark \cite{dnd}. 
We compare our approach with existing supervised methods based on paired images and the latest unsupervised methods based on unpaired images quantitatively and qualitatively.

\begin{table*}[tp]
\begin{center} 
\fontsize{9}{9}\selectfont
\caption{Deraining results of several competitive methods on Rain100L, RealRainL, and Rain12.}
\label{deraining_quantitative_result}
\renewcommand{\arraystretch}{1.6}
\setlength{\tabcolsep}{2.2mm}{
\begin{tabular}{clccccccc}
\toprule
&\multirow{2}[3]{*}{Methods} & GAN-based &\multicolumn{2}{c}{Rain100L} & \multicolumn{2}{c}{RealRainL} & \multicolumn{2}{c}{Rain12} \\
\cmidrule{4-9} & & /Publication & PSNR$\uparrow$ & SSIM$\uparrow$ & PSNR$\uparrow$ & SSIM$\uparrow$ & PSNR$\uparrow$ & SSIM$\uparrow$\\
\midrule
\multirow{2}{*}{Model-based methods} & DSC
 \cite{5} & No/TIP 2007& 27.34 & 0.8490 & 27.76  & 0.8750 &  $-$ & $-$ \\
& GMM \cite{12} & No/CVPR 2014& 29.05 & 0.8720 & 28.87 & 0.9259 & $-$ & $-$ \\
\midrule
\multirow{8}{*}{Supervised} & DDN \cite{ddn} & No/CVPR 2017& 32.38 & 0.9260 & 31.18 & 0.9172 & 34.04 & 0.9330 \\
& RESCAN \cite{rescan} & No/ECCV 2018 & 38.52 & 0.9810 & 31.33 & 0.9261 &  $-$ & $-$ \\
& SPA-Net \cite{spanet} & No/CVPR 2018 & 31.95 & 0.9190 & 30.43 & 0.9470 &  $-$ & $-$ \\
& MSPFN \cite{mspfn} & No/CVPR 2020 & 32.40 & 0.9330 & 35.51 & 0.9670 & $-$ & $-$ \\
& NAFNet \cite{naf} & No/ECCV 2022 & 37.00 & 0.9780 & 38.80 & 0.9860 & 34.81 & 0.9430 \\
& Restormer \cite{restormer} & No/CVPR 2022 & 37.57 & 0.9740 & 40.90 & 0.9850 & $-$ & $-$ \\
& PromptIR \cite{potlapalli2023promptir} & NeurIPS 2024 & 38.34 &  0.9830 & 36.99 &  0.9730 &  35.09 & 0.9450\\
& NeRD-Rain-S \cite{chen2024bidirectional} & CVPR 2024 & 42.00 & 0.9900 & 38.64 & 0.9790 & 35.39 & 0.9420\\
\midrule
\multirow{8}{*}{Unsupervised} & CycleGAN \cite{patch} & Yes/ICCV 2017& 24.61 & 0.8340 & 20.19 & 0.8198 & 21.56 & 0.8450 \\
& NLCL \cite{nlcl} & Yes/CVPR 2022 & 20.50 & 0.7190 & 23.06 & 0.8320 &  22.68 & 0.7350 \\
& DerainCycleGAN \cite{deraincyclegan} & Yes/TIP 2021 & 31.49 & 0.9360 & 28.16 & 0.9010 & 33.52 & 0.9400 \\
& DGP-Cyc-GAN \cite{yasarla2022unsupervised} & Yes/ICPR 2022 & 31.88 & 0.9394 & 29.01 & 0.9195 & 32.03 & 0.9281 \\
& DCDGAN \cite{pjs} & Yes/CVPR 2022&31.82 & 0.9410 & 30.49 & 0.9390 & 31.56 & 0.9240 \\
& DCDGAN \cite{pjs} + ConvIR \cite{convir} & Yes/TPAMI 2024 & 32.56 & 0.9547 & 31.23 & 0.9423 & 32.15 & 0.9301 \\
& DCDGAN \cite{pjs} + NeRD \cite{nerd} & Yes/CVPR 2024&32.34 & 0.9606 & 31.41 & 0.9435 & 32.36 & 0.9297 \\
& RSCP$^{2}$GAN (Ours) & Yes/$-$ & 34.01 & 0.9606 & 32.66 & 0.9460 & 34.24 & 0.9465 \\
\bottomrule
\end{tabular}}%
\end{center}
\end{table*}%

\begin{figure*}[t]
\centering
\includegraphics[width=1\linewidth]{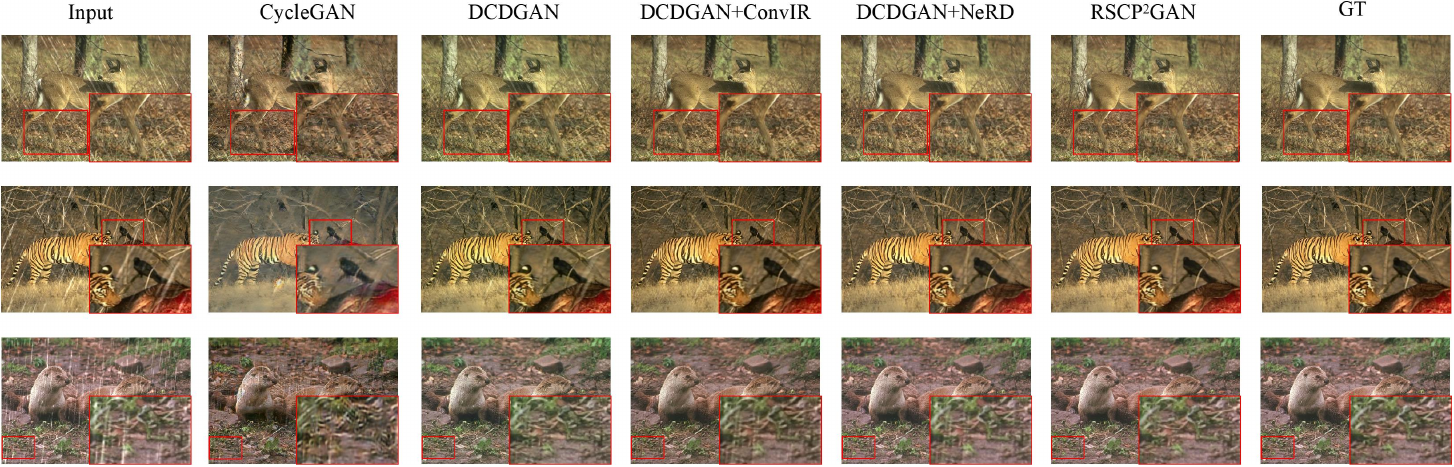}
\caption{Restoration results of our model against other competing methods on the Rain100L \cite{rain100} dataset.}
\label{quyuduibi_1}
\end{figure*}

\begin{figure*}[t]
\centering
\includegraphics[width=1\linewidth]{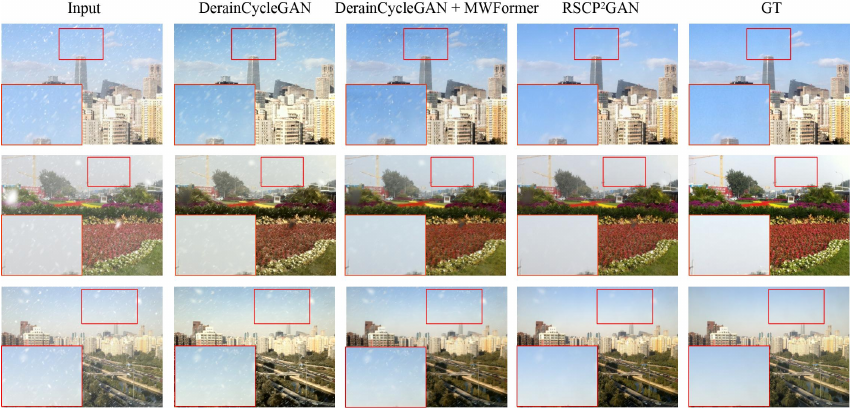}
\caption{Restoration results of our model against other competing methods on the CSD \cite{csd} dataset.}
\label{quxueduibi}
\end{figure*}

\begin{figure}[h]
\centering
\includegraphics[width=0.9\linewidth]{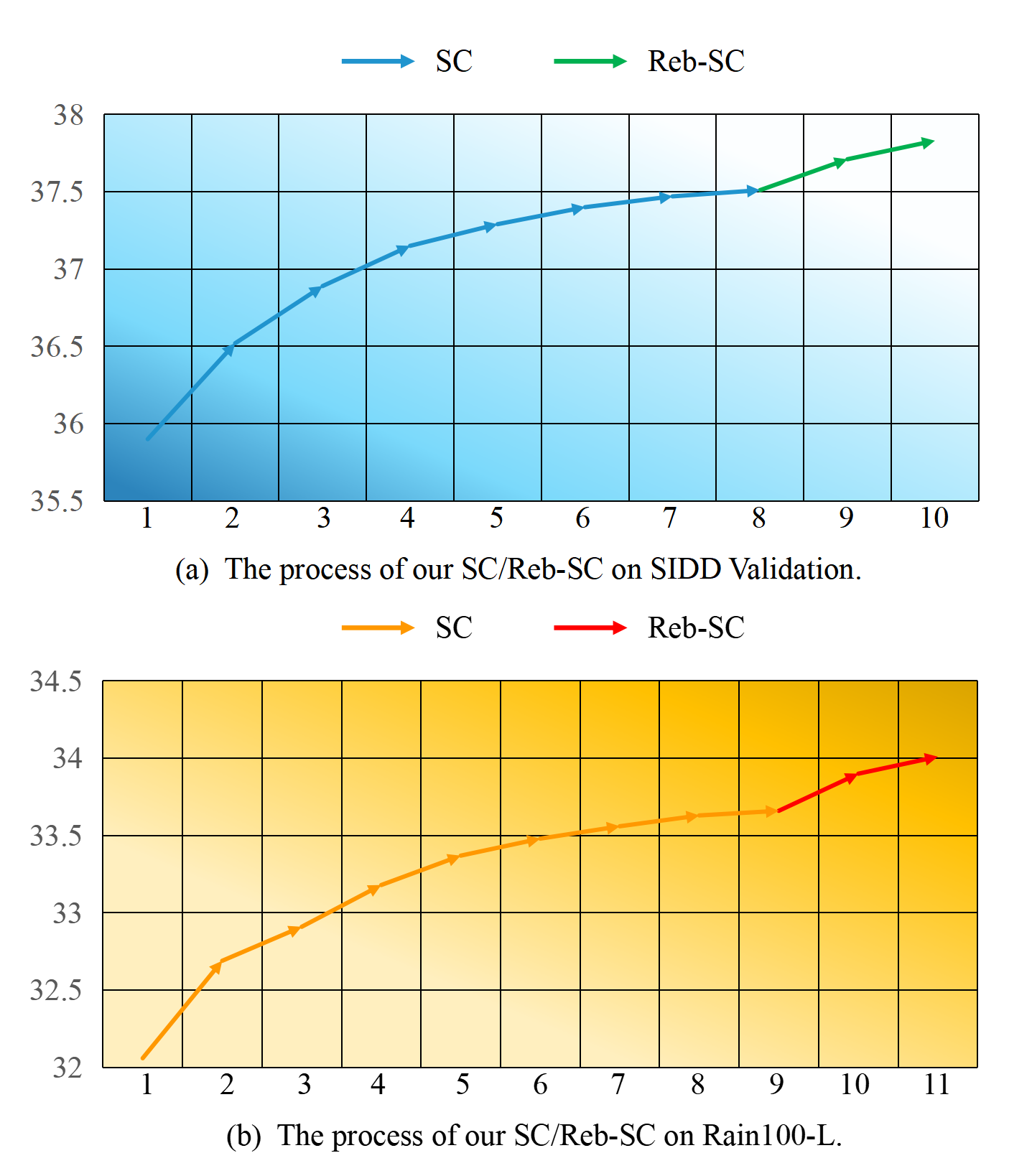}
\caption{
The process of SC and Reb-SC on SIDD Validation \cite{sidd} and Rain100L \cite{rain100}. 
In the SC stage, they need nearly eight iterations. In Reb-SC stage, two iterations are required: the first with the number of augmented images N=2 and the second with N=4.}
\label{ablation_sc}
\end{figure}

\noindent \textbf{Quantitative Comparison.}
We utilize full-reference metrics (PSNR and SSIM) to assess the effectiveness of our method.
Table \ref{denoising_quantitative} shows that our RSCP$^{2}$GAN performs favorably against state-of-the-art methods. 
Compared to single-image self-supervised methods like BNN-LAN \cite{bnnlan} and SDAP (E) \cite{pyz}, RSCP$^{2}$GAN provides a PSNR gain of 0.44 dB and 0.53 dB, and an SSIM gain of 0.024 and 0.013 on the SIDD Benchmark. 
Additionally, compared to recent methods like PUCA \cite{puca} and Complementary-BSN \cite{tcsvt_denoising}, our model achieves improvements of 0.34 dB/0.32 dB on the SIDD Validation set and 0.15 dB/0.26 dB on the SIDD Benchmark. 
Note that the results on the DND \cite{dnd} dataset by PUCA \cite{puca} are obtained from training on the DND Benchmark, unlike other methods trained on the SIDD dataset.
In terms of two-stage GAN-based denoising approaches, RSCP$^{2}$GAN outperforms synthetic pairs methods with self-ensemble (e.g., C2N+DnCNN, C2N+DIDN) on both SIDD and DND datasets. 
Although our method's denoising performance is not as good as some of the latest supervised methods \cite{74, restormer} using real image pairs, these methods require a large number of paired images.
Consequently, they may not be effectively applied to real-world image denoising tasks with insufficient paired images for training.
In contrast, RSCP$^{2}$GAN performs well without paired images, making it suitable for a range of real-world denoising scenarios. 
Our method also achieves favorable results on the PolyU \cite{polyu} dataset, as shown in Table \ref{polyu}.

\begin{table*}[tp]
\begin{center} 
\fontsize{9}{9}\selectfont
\caption{Desnowing results of several competitive methods on CSD and Snow100K.}
\label{desnowing_quantitative_result}
\renewcommand{\arraystretch}{1.6}
\setlength{\tabcolsep}{2.2mm}{
\begin{tabular}{clccccc}
\toprule
&\multirow{2}[3]{*}{Methods} & GAN-based &\multicolumn{2}{c}{CSD} & \multicolumn{2}{c}{Snow100K-S} \\
\cmidrule{4-7} & & /Publication & PSNR$\uparrow$ & SSIM$\uparrow$ &  PSNR$\uparrow$ & SSIM$\uparrow$\\
\midrule
\multirow{7}{*}{Supervised} & MGF \cite{mgf} & No/ICNIP 2013& 13.98 & 0.6700 & 
24.32 & 0.7700 \\
& DesnowNet \cite{snow100k} & No/TIP 2018 & 20.13 & 0.8100 & 32.33 & 0.9500\\
& JSTASR \cite{jstasr} & No/ECCV 2020 & 27.96 & 0.8800 & 31.40 & 0.9012 \\
& InvDSNet \cite{invdsnet} & No/TCSVT 2023 & 31.85 & 0.9600 & 34.39 & $-$ \\
& LMQFormer \cite{lmqformer} & No/TCSVT 2023 & 32.64 & 0.9630 & 34.19 & 0.9436 \\
& PEUNet \cite{peunet} & No/TCSVT 2025 & 38.27 & 0.9900 & 36.76 & $-$ \\
& ConvIR \cite{convir} & No/TPAMI 2024 & 39.10 & 0.9900 & 36.37 & 0.9703\\
\midrule
\multirow{6}{*}{Unsupervised} & CycleGAN \cite{patch} & Yes/CVPR 2020 &16.72&0.6975&20.71&0.7639 \\
& DerainCycleGAN \cite{deraincyclegan} & Yes/TIP 2021 & 20.67&0.7963&22.79&0.8024\\
& DGP-Cyc-GAN \cite{yasarla2022unsupervised} & Yes/ICPR 2022 & 21.35 & 0.8042 & 23.41 & 0.8072 \\
& DerainCycleGAN + ConvIR \cite{convir} & Yes/TPAMI 2024 & 22.18&0.8132&23.97&0.8139 \\
& DerainCycleGAN + MWFormer  \cite{mwformer} & Yes/CVPR 2024 & 22.42	& 0.8179& 24.45& 0.8253 \\
& RSCP$^{2}$GAN (Ours) & Yes/$-$ & 23.38 & 0.8462 & 25.97 & 0.8398  \\
\bottomrule
\end{tabular}}%
\end{center}
\end{table*}%

\begin{table}
\begin{center} 
\fontsize{8}{8}\selectfont
\caption{Ablation studies on the proposed modules. $\textbf{V1}$: (U) Conventional GAN-based unsupervised denoising network only with unpaired synthesis; $\textbf{V2}$: V1 + BGMloss; $\textbf{V3}$: V1 + BGMloss + PL module; $\textbf{V4}$: (S) SGabNet (V1 + BGMloss + PL module + self-synthesis) $\textbf{V5}$: (P) P$^{2}$GAN (our baseline).} 
\label{ablation} 
\setlength{\tabcolsep}{3.0mm}{
\begin{tabular}{ccccccccc}
\toprule
\myrowcolour%
Methods & V1 & V2 & V3 & V4 & V5(ours) \\
\midrule
U & \checkmark & \checkmark & \checkmark & \checkmark & \checkmark \\
BGMloss & & \checkmark &\checkmark & \checkmark & \checkmark \\
PL module & & & \checkmark & \checkmark & \checkmark \\
S & & & & \checkmark & \\
P & & & & & \checkmark \\
\midrule
PSNR(dB) & 34.52 & 34.69 & 34.92 & 35.37 & 35.90 \\
\bottomrule 
\end{tabular}}%
\label{table1}
\end{center} 
\end{table}%

\begin{table*}
\begin{center} 
\fontsize{11}{11}\selectfont
\caption{Ablation studies of Re-boosting module on SC strategy (Reb-SC) on SIDD Validation, SIDD Benchmark, Rain100L, and RealRainL. The N is the number of augmentation images.} 
\renewcommand{\arraystretch}{1.0}
\setlength{\tabcolsep}{3mm}{
\begin{tabular}{ccccccc}
\toprule
\myrowcolour%
 & \textit{N} & SIDD Validation & SIDD Benchmark & Rain100L & RealRainL & Average improve\\
\midrule
\multirow{4}{*}{Baseline} & 0 & 37.51 & 37.43 & 33.66 & 32.28 & 0\\
 & 2 & 37.71 & 37.60 & 33.90 & 32.53 & 0.22\\
 & 4 & 37.83 & 37.69 & 34.01 & 32.66 & 0.33\\
 & 8 & 37.78 & 37.69 & 33.93 & 32.67 & 0.29\\
\midrule
\end{tabular}}%
\label{ablation_reb}
\end{center} 
\end{table*}%

\noindent \textbf{Qualitative Comparison.} Fig. \ref{quzaoduibi} shows that RSCP$^{2}$GAN generates visually pleasing results in terms of detail, color, and naturalness. 
Existing methods often fail to recover image details, over-smooth the noisy images, or generate results with chromatic aberration. 
For example, ASPSN \cite{apbsn}, CVF-SID \cite{cvf}, and LGBPN \cite{lgbpn} over-smooth images and generate results without details. 
BNN-LAN \cite{bnnlan} and C2N \cite{jang} may cause image blurring, while CBDNet \cite{25} sometimes results in chromatic aberration. 
In contrast, RSCP$^{2}$GAN better removes noise, preserves details, and avoids chromatic aberration. 
The results on the PolyU dataset, shown in Fig. \ref{quzaoduibi_polyu}, demonstrate that our method effectively preserves details that other methods may mistakenly remove.

\subsection{Image Deraining}\label{sec4_3}
We evaluate the proposed method and state-of-the-art approaches on image draining benchmark datasets.  

\noindent \textbf{Quantitative Comparison.} We evaluate the deraining performance of our method on the Rain100L \cite{rain100}, RealRainL \cite{realrain} and Rain12 \cite{rain12} datasets. 
We note that there are few unsupervised deraining methods with source codes for performance evaluation. Besides the existing unsupervised deraining frameworks, DerainingCycleGAN \cite{deraincyclegan}, NLCL \cite{nlcl}, and DCDGAN \cite{pjs}, we add two new deraining networks, ConvIR \cite{convir} (TPAMI2024) and NeRD \cite{nerd} (CVPR2024) as comparison methods. For fair unsupervised comparison, these methods are incorporated into the DCDGAN framework as the restorer, and are trained using the same datasets as ours.
Table \ref{deraining_quantitative_result} shows that RSCP$^{2}$GAN outperforms all unsupervised approaches. 
Compared to DerainCycleGAN \cite{deraincyclegan} and DCDGAN \cite{pjs}, our method achieves a PSNR gain of 2.52 dB and 2.19 dB, and an SSIM gain of 0.024 and 0.019 on the Rain100L test set. 
Our method also performs comparably to some supervised methods like SPA-Net \cite{spanet} and DDN \cite{ddn}. Equipped with more advanced restorers, both DCDGAN+ConvIR and DCDGAN+NeRD achieve better performance than the original DCDGAN. However, our RSCP$^{2}$GAN (using the traditional NAFNet as the restorer) still achieves the best results in three test sets. This indicates that the improvement from superior restorers is limited compared to advancements in restoration mechanisms, which further highlights the effectiveness of our SC and Reb-SC strategies.

\noindent \textbf{Qualitative Comparison.} Fig. \ref{quyuduibi_1} visually compares deraining methods on the Rain100L dataset. As shown, traditional methods such as CycleGAN and DCDGAN struggle to recover fine textures and complex degradations, often leaving visible artifacts, over-smoothed regions, and broken structures. The introduction of structural enhancement modules like ConvIR and NeRD leads to moderate improvements in the DCDGAN variants, resulting in clearer local details and slightly improved structural consistency. However, these methods still exhibit texture distortion and edge instability, particularly in scenes with dense textures or complex backgrounds.
In contrast, RSCP$^{2}$GAN consistently delivers sharper and more natural restoration results across all samples. It better preserves detailed textures, such as animal fur and background foliage, and maintains strong edge continuity, effectively suppressing artifacts and avoiding over-smoothing. In the zoomed-in regions, RSCP²GAN's outputs closely resemble the ground truth (GT), demonstrating superior capability in both degradation modeling and restoration learning. These observations confirm the robustness and generalization ability of our framework in real-world restoration scenarios.

\subsection{Image Desnowing}\label{sec4_5}
We evaluate the proposed method and state-of-the-art approaches on image desnowing benchmark datasets. The results of the supervised methods from previous works are presented as a reference. The comparison focuses on the performance of unsupervised methods.

\noindent \textbf{Quantitative Comparison.} Since there is limited exploration of unsupervised desnowing in existing work, we adopt the classic unsupervised restoration architecture, DerainCycleGAN, as a baseline for comparison, and replace its restoration module with the latest restoration networks, ConvIR \cite{convir} and MWFormer \cite{mwformer}. Unsupervised training consistent with the proposed method is conducted using the aforementioned desnowing dataset. In contrast, our framework still uses traditional NAFNet as the restorer, which is consistent with the deraining task. As shown in the Table \ref{desnowing_quantitative_result}, for the desnowing task, despite not using the latest restoration networks, our method still achieves a significant advantage on both test sets, thanks to the design of our RSCP$^{2}$GAN framework. Specifically, compared to DerainCycleGAN \cite{deraincyclegan}, DGP-Cyc-GAN \cite{yasarla2022unsupervised}, and DerainCycleGAN + MWFormer \cite{mwformer}, our model provides a PSNR gain of 2.71 dB, 2.03 dB, and 0.96 dB, and an SSIM gain of 0.050, 0.042, and 0.028 on the CSD test set.

\noindent \textbf{Qualitative Comparison.} The Fig. \ref{quxueduibi} presents a visual comparison between existing methods and our approach on the CSD dataset. Our method can effectively remove snowflakes, while others tend to leave residual snow. Additionally, existing methods often produce blurry results in snow removal scenarios with complex textures, whereas our approach preserves more accurate and detailed textures.

\subsection{Ablation Study}
\label{sec4_4}
\noindent \textbf{Effectiveness of the Proposed Framework.} We validate the effectiveness of the P$^{2}$GAN structure as described in Table \ref{ablation} for the denoising task. 
Here, V1 represents a GAN-based unsupervised denoising network with only unpaired synthesis. 
In addition, V2 extends V1 by adding the BGM loss, while V3 further includes the PL module. V4 introduces the branch U-S, and V5 is our baseline (P$^{2}$GAN).

Table \ref{ablation} shows that adding the BGM loss to the GAN-based unsupervised network (V1 to V2) results in a 0.17 dB PSNR improvement.
Incorporating the PL module (V2 to V3) yields an additional PSNR increase of approximately 0.23 dB, highlighting the effectiveness of the PL module in enhancing synthetic image quality. When adding the ``self-synthesis'' constraint to V3 to obtain V4, there is a notable improvement of 0.45 dB PSNR on the SIDD Benchmark, indicating that combining ``self-synthesis'' with unpaired synthesis enhances network training and restorer performance. 
Comparing P$^{2}$GAN with V4, we observe a substantial performance boost in P$^{2}$GAN, with PSNR gains of 0.53 dB on the SIDD Validation. This demonstrates that P$^{2}$GAN produces more realistic synthetic degraded images, improving the restorer’s performance.

\noindent \textbf{Effectiveness of the SC and Reb-SC Strategies.} We apply the SC and Reb-SC strategies to our baseline method (P$^{2}$GAN), resulting in SCP$^{2}$GAN and RSCP$^{2}$GAN. 
Fig. \ref{ablation_sc} shows the performance of the restorer after each iteration on the SIDD Validation and Rain100L datasets. SCP$^{2}$GAN demonstrates significant improvement in the initial iterations, with gains exceeding 0.5 dB on both datasets in the first iteration. 
However, the PSNR improvement between subsequent iterations decreases, with the final iteration showing only about a 0.03 dB gain. For RSCP$^{2}$GAN, after the SC strategy converges, the Reb-SC strategy results in a substantial improvement of approximately 0.3 dB. 
Compared to P$^{2}$GAN, RSCP$^{2}$GAN achieves significant improvements of 1.93 dB on the SIDD Validation and 1.95 dB on Rain100L. 
This demonstrates that our approach achieves state-of-the-art performance in image denoising and deraining and shows the general applicability of the SC strategy.

\noindent \textbf{Effectiveness of the Number of Augmentation Images (N) in the Re-boosting (ReB) Module.} Table \ref{ablation_reb} shows that Reb-SC improves performance across all datasets. 
For $N$=2, the improvement is approximately 0.2 dB; for $N$=4, it is about 0.3 dB. However, when $N$=8, the performance improvement is less significant than $N$=4.

\begin{figure}[t]
\centering
\includegraphics[width=1\linewidth]{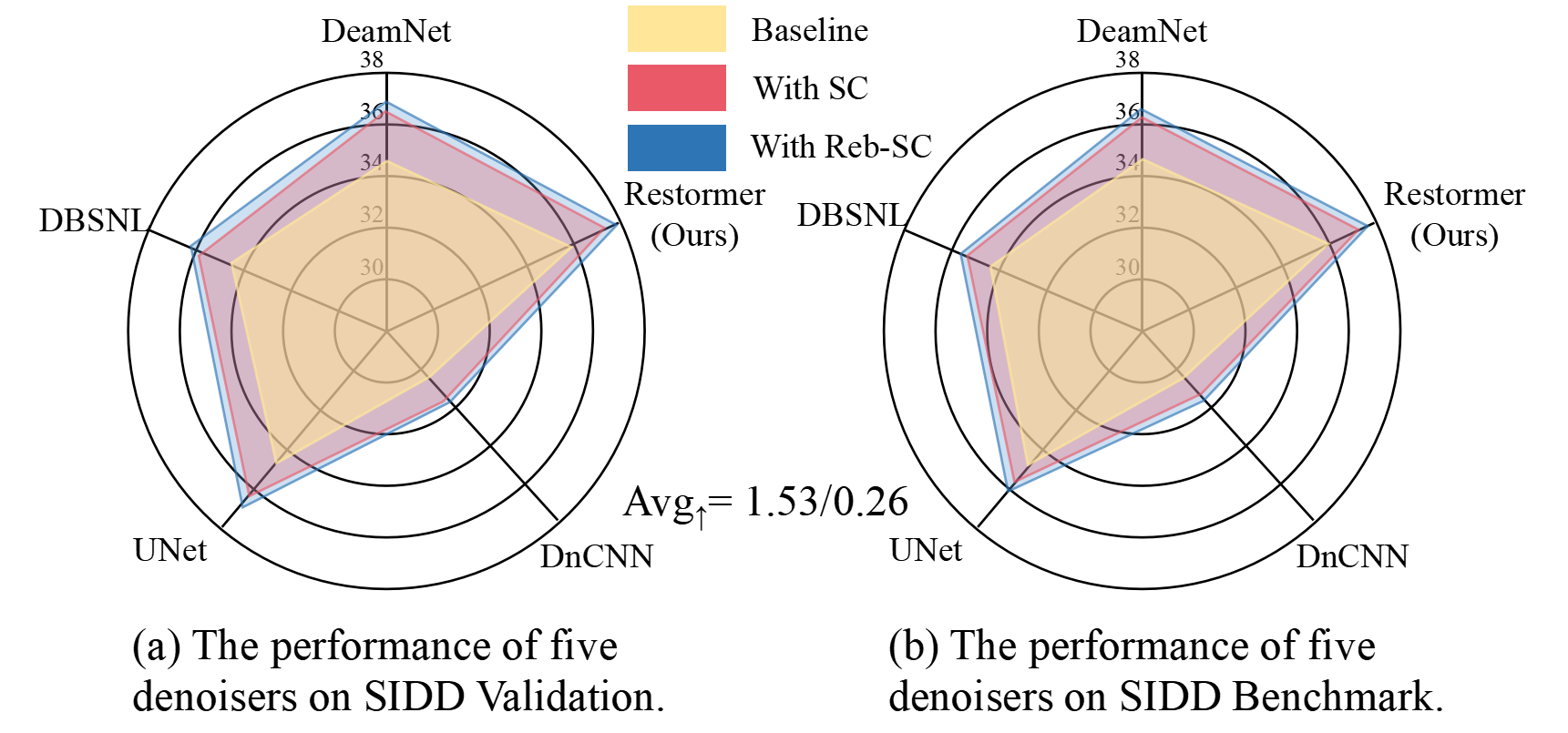}
\caption{Effectiveness of SC and Reb-SC strategy on five restorers.}
\label{trans}
\end{figure}

\subsection{Analysis on Transferability}\label{sec4_5} To evaluate the transferability of our SC and Reb-SC strategies, we apply them to various classical and modern restorers, including DnCNN \cite{17}, UNet \cite{unet}, DeamNet \cite{aind1}, and DBSNL \cite{dbsnl}. 
The framework still adopts the P$^{2}$GAN proposed in section \ref{sec3} with different restorers.
Fig. \ref{trans} shows that the SC and Reb-SC strategies are effective across these networks.
For example, applying SC to DnCNN yields PSNR/SSIM improvements of 0.96 dB/0.0085 on the SIDD Validation dataset. 
For UNet and DBSNL, the gains are 1.58 dB/0.0373 and 1.52 dB/0.0035, respectively. Reb-SC also shows improvement across multiple restorers, demonstrating its high transferability and potential applicability to other restorers within unsupervised image restoration frameworks.

\section{Conclusion}\label{sec5} In this paper, we first introduce Parallel Prompt GAN (P$^{2}$GAN) for unsupervised image restoration as our baseline. 
Furthermore, we propose an SC strategy to provide the $Res$ and PL modules with a self-boosting capacity and significantly improve restoration performance. 
To improve the performance of the $Res$, we apply a Reb-SC strategy, which leads to further enhancement of the $Res$ module using the SC strategy. 
Extensive experimental results show that the proposed method achieves state-of-the-art performance. 
In addition, We also demonstrate the transferability of the SC and Reb-SC strategies to various restorers, indicating their broad applicability to low-level computer vision tasks.


Although the SC and Reb-SC strategies proposed in this paper significantly improve denoising and deraining performance within an unsupervised GAN framework, these are aimed at individual restoration tasks.
Real-world scenarios often involve mixed degradations, such as low resolution, motion blur, adverse weather conditions, and compression artifacts. 
Future work will explore these more complex scenarios and evaluate the generalization capabilities of the SC and Reb-SC strategies across diverse restoration challenges.

{\small
\bibliographystyle{IEEEtran}
\bibliography{e}
}

\newpage

\section{Biography Section}

\begin{IEEEbiography}
	[{\includegraphics[width=1in,height=1.25in,clip,keepaspectratio]
		{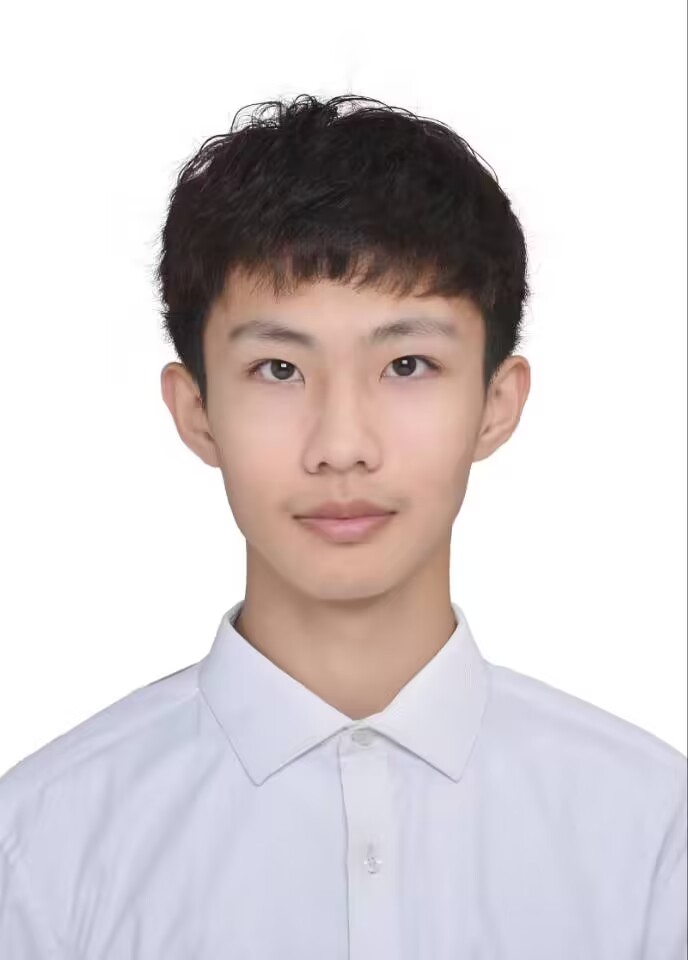}}]
	{Xin Lin} is currently pursuing a B.Eng. degree in electronics and information engineering from Sichuan University, Chengdu, China, under the supervision of Prof. Chao Ren. Now, he is a Visiting Student at the University of California, Merced, under the supervision of Prof. Ming-Hsuan Yang. He serves as a reviewer of some top-tier conferences and journals such as the CVPR, NeurIPS, ECCV, ACM MM, IJCV, and Information Fusion. His research interests include 3D Vision and image restoration.
\end{IEEEbiography}

\begin{IEEEbiography}
	[{\includegraphics[width=1in,height=1.25in,clip,keepaspectratio]
		{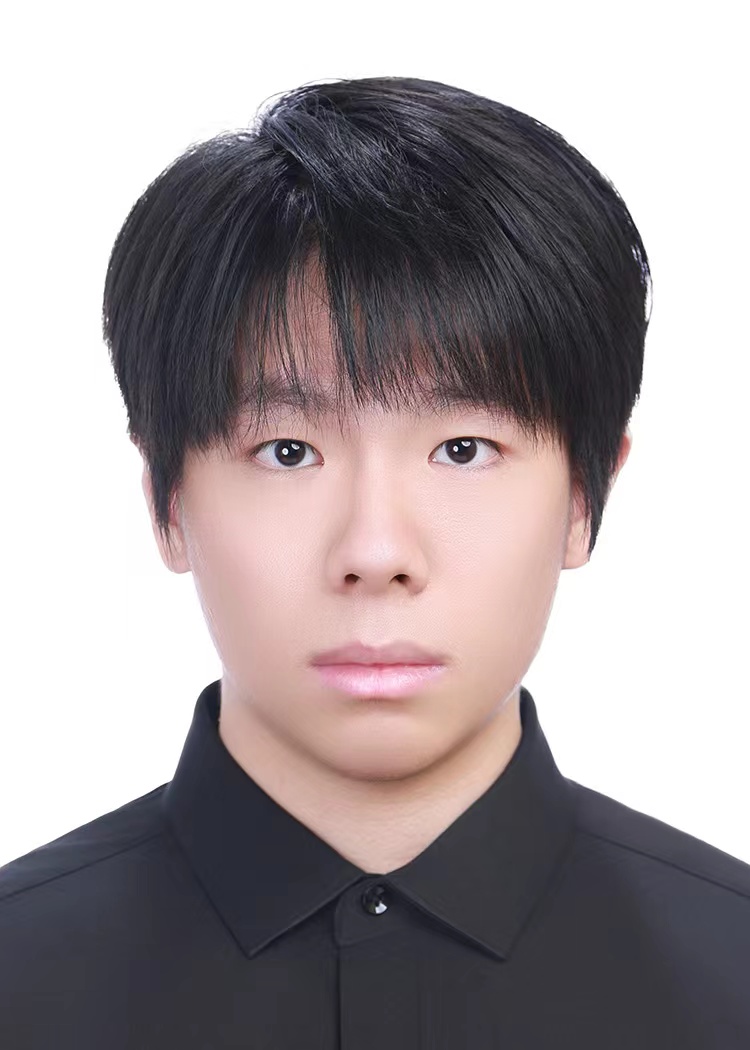}}]
	{Yuyan Zhou} is a Ph.D. student in the Department of Computer Science      and Engineering of Hong Kong University of Science and
        Technology (HKUST). He received his B.Eng degree in Artificial Intelligence from Nanjing University of Aeronautics and Astronautics in 2024. He served as a reviewer of some top-tier conferences and journals such as the CVPR, NeurIPS, and IJCV. His research interests include machine learning and its application on computer vision and natural languague processing.
\end{IEEEbiography}

\begin{IEEEbiography}
	[{\includegraphics[width=1in,height=1.25in,clip,keepaspectratio]
		{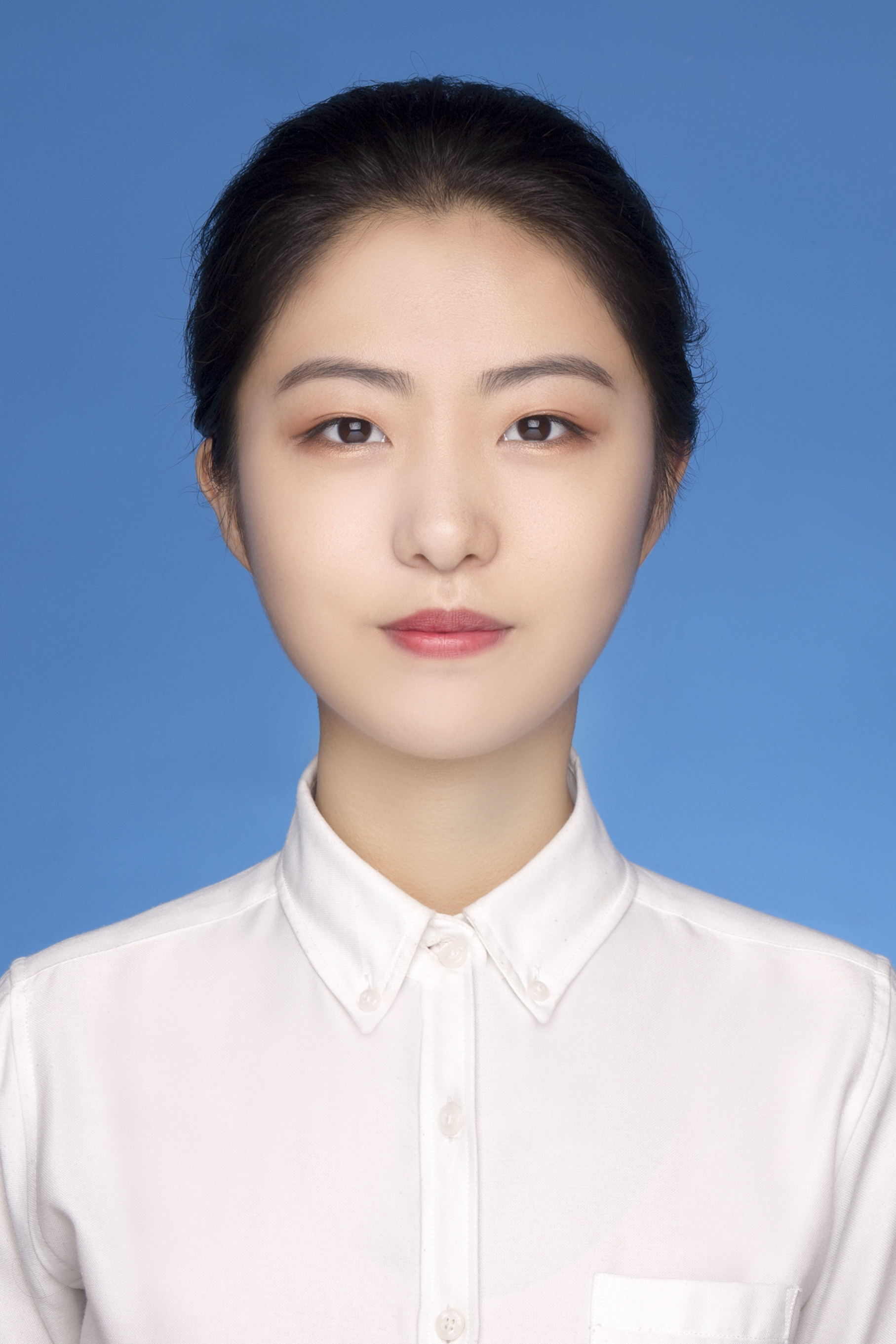}}]
	{Jingtong Yue} is currently pursuing a B.Eng. degree in Electronics and Information Engineering at Sichuan University in Chengdu, China, under the supervision of Professor Chao Ren. Now she is also involved in research on intelligent robots in space at the Institute of Space and Astronautics, Sichuan University. Her research interests include radar point cloud processing, image restoration, and multi-modal fusion.
\end{IEEEbiography}

\begin{IEEEbiography}
	[{\includegraphics[width=1in,height=1.25in,clip,keepaspectratio]
		{figure/ChaoRen.pdf}}]
	{Chao Ren} (Member, IEEE) received the B.S.degree in electronics and information engineering and Ph.D degree in communication and information system from Sichuan University, Chengdu, China, in 2012 and 2017, respectively. From 2015 to 2016, he was a Visiting Scholar with the Department of Electrical and Computer Engineering, University of California at San Diego, La Jolla, CA, USA. He is currently an Associate Research Professor at Sichuan University College of Electronics and Information Engineering. His research interests include inverse problems in image and video processing. He has authored more than 60 papers in journals/conferences such as TIP, CVPR, ICCV, ECCV, NeurIPS, etc. He has won the Huawei Spark Award and was selected for the National Post-Doctoral Program for Innovative Talents of China.

\end{IEEEbiography}

\begin{IEEEbiography}
	[{\includegraphics[width=1in,height=1.25in,clip,keepaspectratio]
		{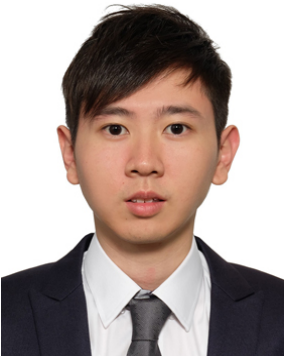}}]
	{Kelvin C. K. Chan} is a research scientist at Google DeepMind. Before joining Google, he was a Ph.D. student at MMLab@NTU under the supervision of Prof. Chen Change Loy. He received his M.Phil. degree in mathematics as well as his B.Sc. and B.Eng. degrees from The Chinese University of Hong Kong. His current research interest focuses on low-level vision and multimodal content generation.

\end{IEEEbiography}

\begin{IEEEbiography}
	[{\includegraphics[width=1in,height=1.25in,clip,keepaspectratio]
		{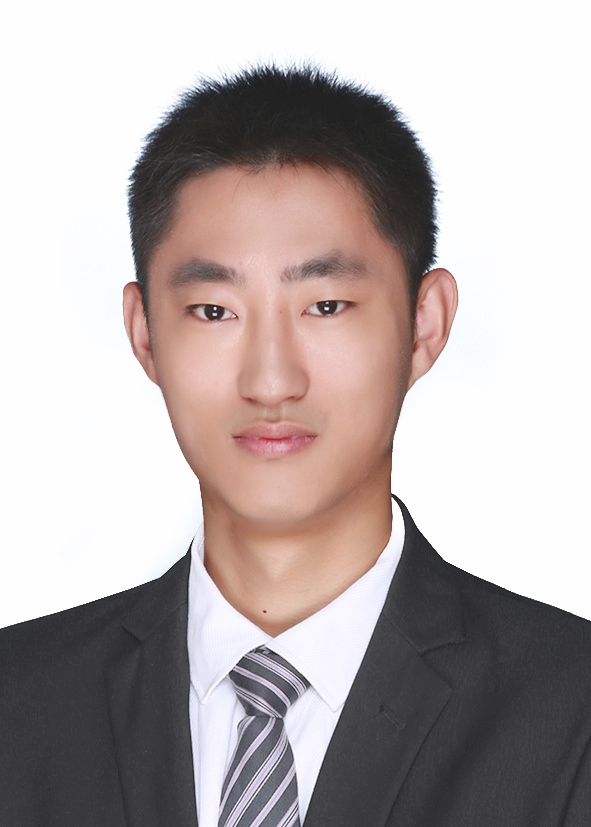}}]
	{Lu Qi} works as a postdoc with Prof. Ming-Hsuan Yang at UC Merced and has about 8,000 citations in Google Scholar. He received his Ph.D. degree from The Chinese University of Hong Kong in 2021 and obtained the Hong Kong Ph.D. Fellowship in 2017. His research interests include instance-level detection, image generation, and cross-modal pretraining. He was the senior program chair of AAAI 2023/2024 and area chair of NeurlPS2024 and ICLR2024.
\end{IEEEbiography}

\begin{IEEEbiography}
[{\includegraphics[width=1in,height=1.25in,clip,keepaspectratio]
	{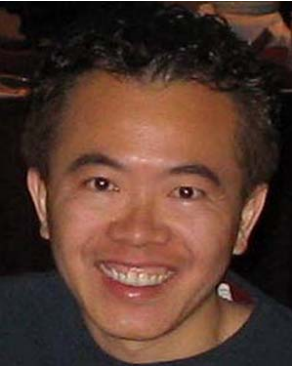}}]
{Ming-Hsuan Yang} is a professor of Electrical Engineering and Computer Science at the University of California, Merced. Yang serves as a program co-chair of the IEEE International Conference on Computer Vision (ICCV) in 2019, program co-chair of the Asian Conference on Computer Vision (ACCV) in 2014, and general co-chair of ACCV 2016. Yang currently serves as an associate editor-in-chief of IEEE Transactions on Pattern Analysis and Machine Intelligence (PAMI) and as an associate editor of the International Journal of Computer Vision (IJCV), Image and Vision Computing, and Journal of Artificial Intelligence Research. He received the NSF CAREER award in 2012 and Google Faculty Award in 2009. He is a Fellow of the IEEE and ACM.
\end{IEEEbiography}

\end{document}